\documentclass[lettersize,journal]{IEEEtran}
\usepackage{amsmath,amsfonts}
\usepackage{algorithmic}
\usepackage{algorithm}
\usepackage{array}
\usepackage{textcomp}
\usepackage{stfloats}
\usepackage{url}
\usepackage{verbatim}
\usepackage{graphicx}
\usepackage{cite}
\usepackage{url}            
\usepackage{booktabs}       
\usepackage{amsfonts}       
\usepackage{nicefrac}       
\usepackage{microtype}      
\usepackage[table]{xcolor}         

\usepackage{algorithm}
\usepackage{algorithmic}
\usepackage{amsmath}
\usepackage{amssymb}
\usepackage{color}
\usepackage{multirow}
\usepackage{subfigure}
\usepackage{bbding}
\usepackage{longtable}
\usepackage{supertabular}
\usepackage{graphicx}
\usepackage{multirow}
\usepackage{wrapfig}
\usepackage{xspace}

\usepackage{enumitem}
\usepackage{pifont}
\usepackage{CJK}
\usepackage{makecell}

\usepackage[pagebackref,breaklinks,colorlinks]{hyperref}
\newcommand{\ie}{\emph{i.e.}\xspace}
\newcommand{\eg}{\emph{e.g.}\xspace}

\begin{document}

\title{Parameter-Inverted Image Pyramid Networks for Visual Perception and Multimodal Understanding}

\author{Zhaokai Wang, Xizhou Zhu, Xue Yang, Gen Luo, Hao Li, Changyao Tian, Wenhan Dou, Junqi Ge, \\Lewei Lu, Yu Qiao, Jifeng Dai
\thanks{
Zhaokai Wang and Xue Yang are with Shanghai Jiao Tong University. 

Xizhou Zhu, Wenhan Dou, Junqi Ge, and Jifeng Dai are with Tsinghua University. 

Gen Luo and Yu Qiao are with Shanghai Artificial Intelligence Laboratory. 

Hao Li and Changyao Tian are with The Chinese University of Hong Kong. 

Lewei Lu is with Sensetime. 

Zhaokai Wang, Xue Yang, Hao Li, Changyao Tian and Jifeng Dai are also with Shanghai Artificial Intelligence Laboratory.

Zhaokai Wang, Xizhou Zhu and Xue Yang contribute equally to this work. 

Corresponding author: Jifeng Dai (daijifeng@tsinghua.edu.cn).

A preliminary version of this research is published in NeurIPS 2024~\cite{piip} as Spotlight.
}
}

\maketitle

\begin{abstract}
Image pyramids are widely adopted in top-performing methods to obtain multi-scale features for precise visual perception and understanding.
However, current image pyramids use the same large-scale model to process multiple resolutions of images, leading to significant computational cost. 
To address this challenge, we propose a novel network architecture, called Parameter-Inverted Image Pyramid Networks (PIIP). 
Specifically, PIIP uses pretrained models (ViTs or CNNs) as branches to process multi-scale images, where images of higher resolutions are processed by smaller network branches to balance computational cost and performance. To integrate information from different spatial scales, we further propose a novel cross-branch feature interaction mechanism. 
To validate PIIP, we apply it to various perception models and a representative multimodal large language model called LLaVA, and conduct extensive experiments on various tasks such as object detection, segmentation, image classification and multimodal understanding. PIIP achieves superior performance compared to single-branch and existing multi-resolution approaches with lower computational cost.
When applied to InternViT-6B, a large-scale vision foundation model,  PIIP can improve its performance by 1\%-2\% on detection and segmentation with only 40\%-60\% of the original computation, finally achieving 60.0 box AP on MS COCO and 59.7 mIoU on ADE20K. For multimodal understanding, our PIIP-LLaVA achieves 73.0\% accuracy on TextVQA and 74.5\% on MMBench with only 2.8M training data.
Our code is released at \url{https://github.com/OpenGVLab/PIIP}.
\end{abstract}


\section{Introduction}
\label{sec:intro}
In modern computer vision, advanced image perception and understanding systems heavily depend on large-scale pretrained models, which often require tens of thousands to millions of GPU hours for pretraining~\cite{deit,deit3,augreg, VLP:CLIP}. To adapt these costly pretrained models for vision perception  (\eg, object detection~\cite{cai2018cascade,deformable_detr,yang2019scrdet,yang2021r3det} and segmentation~\cite{mask_rcnn, upernet, wen2022patchdct}) tasks, the common practice is to combine them with image pyramids~\cite{singh2018sniper, najibi2019autofocus}, \ie upsample and downsample the image to multiple scales and process them independently, then fuse their outputs.
This combination is crucial in constructing multi-scale features for visual perception and understanding on high-resolution images.

\begin{figure*}[t]
    \centering
    \includegraphics[width=0.85\linewidth]{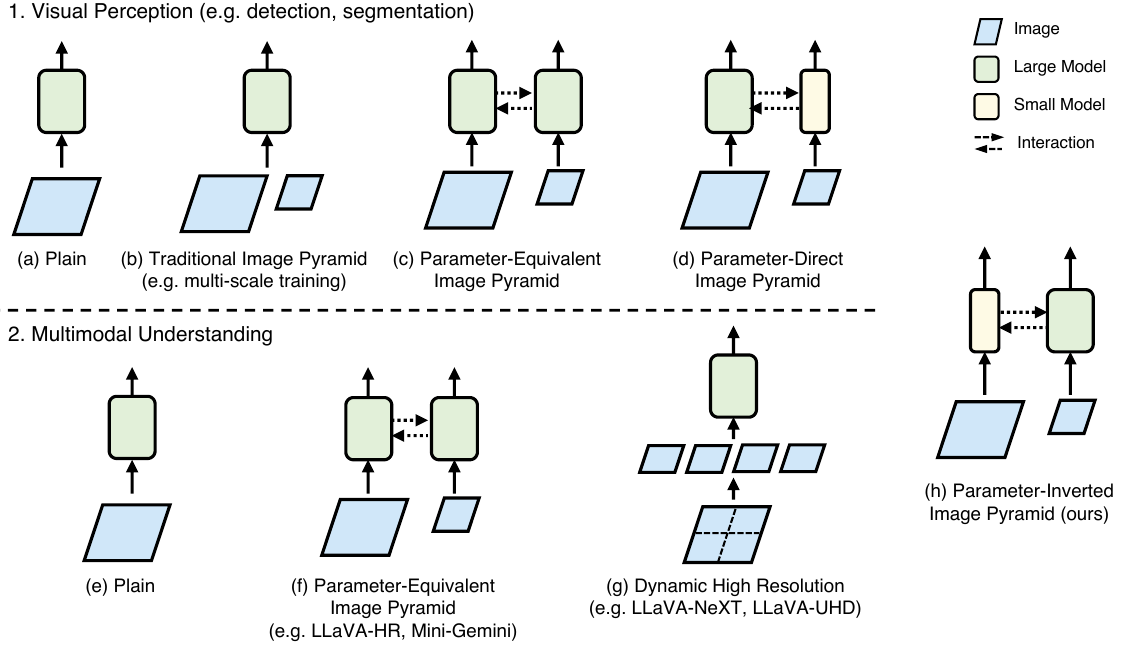}
    \caption{\textbf{Different multi-resolution designs in visual perception and multimodal understanding.} 
    \textbf{(a)(e)} Plain network without multi-scale features. 
    \textbf{(b)(c)(f)} Inefficient image pyramid networks using equivalently large models for all scales, either with shared weights or with separate weights and interactions. 
    \textbf{(d)} Parameter-direct image pyramid network which processes high-resolution images with large models, leading to high computational cost. 
    \textbf{(g)} Multi-resolution approaches on multimodal tasks based on grid partition.
    \textbf{(h)} Our efficient and effective parameter-inverted image pyramid network (PIIP), which pairs models of increasing parameter sizes inversely with images of decreasing resolution. It achieves better performance with much lower computational cost.
    }
    \label{fig:pyramid}
\end{figure*}

Nonetheless, employing image pyramids with pretrained models incurs substantial computational overhead. Current image pyramids use the same large-scale model to process multiple scales of the same image (Figure~\ref{fig:pyramid} (b)), leading to a quadratic growth in computational demand as image resolution scales increase. 
When a certain computation budget is imposed, the maximum image resolution is limited, which negatively impacts model performance in visual perception tasks.
Although feature pyramids~\cite{fpn,nas-fpn, bifpn} aim to reduce this overhead in dense prediction tasks, most top-ranking models~\cite{internimage,eva,codetr,vit_adapter} in the MS COCO challenge~\cite{coco} still rely on image pyramids due to their performance advantages. Therefore, it is necessary to reduce the computing resources needed to build image pyramids while maintaining high performance for visual perception.

High computational costs are also a challenge in multimodal understanding. In the field of multimodal large language models (MLLMs)~\cite{VLM:GPT-4v, VLM:LLaVA-1.5, mono_internvl, internvl2.5, sparkle, gamma_mod, bai2023qwenvl}, previous works have validate that scaling up the resolution of the input image helps to improve the visual understanding ability of MLLMs. 
Some multi-resolution approaches also adopt image pyramids~\cite{llava-hr, VLM:MiniGemini, mg_llava} (Figure~\ref{fig:pyramid} (f)). 
Others adopt a dynamic high resolution strategy, \ie partition the image into multiple slices and pass through the vision encoder~\cite{VLM:LLaVA-NeXT, llava_uhd, llava_uhd_v2} (Figure~\ref{fig:pyramid} (g)).
However, these methods use visual models with same parameter scale to process large scale and small scale inputs. This inevitably imposes great computational burden, limiting the performance of multimodal understanding under computation budget. 
Despite that the vision encoder only take up a small fraction of the total FLOPs compared with the language model, its depth and input size significantly affect the inference speed, as examined in~\cite{mono_internvl, eve}.
Consequently, it is also essential to increase the input resolution while maintaining an acceptable computation cost for multimodal understanding.

To address this issue, our primary insight is that it is unnecessary to utilize vision models of identical size across all resolutions (Figure~\ref{fig:pyramid}(b-c)) or adopt a parameter-direct design (Figure~\ref{fig:pyramid}(d)). Instead, we can adopt a \emph{parameter-inverted} design and allow features at different resolutions to complement each other through effective feature fusion. This approach improves computational efficiency and avoids redundant modeling of similar information.
In lower-resolution pyramid levels, larger models can efficiently extract rich semantic and contextual features because of the smaller input image size. Conversely, high-resolution branches should focus on capturing the missing detail information without reprocessing semantic data. Thus, high-resolution features can focus on smaller receptive fields with less semantic information, reducing computational costs while retaining important details.

Building on this principle, we propose to construct a high-performance, low-cost image pyramid network. This design employs a series of models with increasing parameter sizes paired with decreasing image resolutions, as illustrated in Figure~\ref{fig:pyramid}(h). 
Each resolution level directly leverages existing pretrained vision foundation models for feature extraction, eliminating the need to train multi-scale image pyramid networks from scratch. 
Additionally, strong feature interactions between levels ensure that features at varying scales remain complementary and avoid redundant feature computation.

To implement this, we propose Parameter-Inverted Image Pyramid Networks (PIIP), which leverage the complementary nature of features at different resolutions.  This network processes images of various scales using smaller branches for high-resolution inputs to capture local details and larger branches for low-resolution inputs to extract global context. Each branch is initialized from a pretrained Vision Transformers (ViTs)~\cite{vit} or Convolutional Neural Networks (CNNs)~\cite{convnext}. Additionally, we introduce a feature interaction module that allows features between different resolutions to complement each other. A dedicated feature interaction module facilitates the integration of features across multiple scales, reducing the parameter scale of high-resolution branches and ensuring effective information exchange. This approach significantly lowers computational expenses while maintaining superior performance.

To validate the capabilities of our method on general visual perception and understanding tasks, we evaluate its performance on multimodal understanding, object detection, segmentation and image classification.
Our method outperforms single-branch networks and traditional image pyramids while reducing computational costs across all these tasks. 
PIIP provides a new direction for efficient and effective visual computing. Our contributions are as follows\footnote{
This paper is built upon our work published in NeurIPS 2024 as Spotlight (Top 2.08\%)~\cite{piip} with substantial extensions. Compared to the original version, we extend  PIIP in five aspects in terms of model designs and experiments.

\textbf{1)} We extend PIIP to multimodal understanding and propose a novel multimodal large language model called PIIP-LLaVA. Experiments demonstrate that  our PIIP can significantly help MLLM improve the performance over existing multi-resolution approaches, \eg +1.4\% on average over LLaVA-HR. PIIP-LLaVA achieves 73.0\% on TextVQA and 74.5\% on MMBench with only 2.8M training data (Section~\ref{sec:method_merging}, \ref{sec:exp_multimodal}). 
\textbf{2)} We extend PIIP to different types of visual architectures, including ViT-based architectures and CNN-based architectures. Extensive experiments have validated the generalizability and effectiveness of PIIP (Section~\ref{sec:method_branches}, \ref{sec:exp_det}). 
\textbf{3)} PIIP is also adapted to  heterogeneous ViT-CNN structures, which can effectively leverage  the distinct advantages of  ViT and CNN (Section~\ref{sec:method_branches}, \ref{sec:exp_det}, \ref{sec:exp_multimodal}). 
\textbf{4)} More experiments are conducted on different settings, including from-scratch pretraining and stronger detection models. Based on  InternViT-6B, PIIP finally achieves 60.0 box AP on object detection  (Section~\ref{sec:exp_det}, \ref{sec:exp_cls}). 
\textbf{5)} We provide more ablation studies and qualitative analysis to further understand the design principle and the capability of PIIP (Section~\ref{sec:exp_ablation}, ~\ref{sec:exp_visualization}, \ref{sec:exp_qualitative}).
}:
\begin{itemize}[leftmargin=*]
    \item  We identify the key issue  of common image pyramids in computer vision, \emph{i.e.,} the expensive computational overhead. To overcome this limitation, we introduce a novel framework named Parameter-Inverted Image Pyramid (PIIP) to enhance the multi-scale representational capability of vision backbones while maintaining high computational efficiency.

    \item In PIIP, we propose a novel strategy to promote the information interactions across different pyramid branches, namely cross-branch interaction. It integrates features of different spatial scales and semantic levels for better visual representation learning.

    \item We extend PIIP to various visual architectures, including pure-ViT, pure-CNN and the heterogeneous ViT-CNN networks. The heterogeneous structure leverages pretrained ViTs like CLIP~\cite{VLP:CLIP} for global semantic modeling and CNNs for local feature extraction.

    \item  Based on PIIP, we propose a new MLLM termed as PIIP-LLaVA, which adopts the PIIP design to achieve efficient and effective high-resolution understanding. Beyond visual perception, PIIP-LLaVA demonstrate the versatility of PIIP in multimodal understanding.
    
    \item  We apply our method on various visual perception tasks, including object detection, segmentation, and image classification. Our method surpasses single-branch models and other image pyramid methods with higher performance and lower computation cost.
    When applied to InternViT-6B~\cite{internvl}, a large-scale vision foundation model, PIIP improves its performance on object detection and semantic segmentation by 1.9\% and 1.3\% while reducing 43\% and 58\% of computational costs, respectively. We finally achieve 60.0 $\rm AP^b$ on COCO~\cite{coco} and 59.7 mIoU on ADE20K~\cite{ade20k}. We also provide extensive ablation studies and valuable design guidelines for PIIP that may benefit future research.

    \item We evaluate PIIP-LLaVA on various MLLM benchmarks, which demonstrates superior performance over existing multi-resolution approaches, \eg +1.4\% on average over LLaVA-HR~\cite{llava-hr}, or 73.0\% on TextVQA and 74.5\% on MMBench with only 2.8M training data.  

\end{itemize}

\section{Related Work}

\subsection{Image Pyramids and Feature Pyramids} 
In order to obtain multi-scale image understanding capabilities for visual perception tasks, especially for dense prediction tasks, modern methods widely adopt image pyramids or feature pyramids.
Image pyramids~\cite{snip, singh2018sniper, najibi2019autofocus} resize the original image into different resolutions and feed into the model separately, allowing for accurately detecting objects at various scales. However, this technique introduces significant computational costs for high-resolution images. 
Feature pyramids~\cite{fpn, nas-fpn, bifpn} represent another method for constructing multi-scale feature representations by combining low-resolution, semantically strong features with high-resolution, semantically weak features. Despite significant reduction in computational costs, feature pyramids cannot fully replace image pyramids when detecting very small or large objects, limiting their performance~\cite{snip}. 
Our proposed PIIP adopts the multi-resolution input paradigm from image pyramids while integrating feature interaction from feature pyramids. It introduces the parameter-inverted design to achieve efficient and effective visual perception.

\subsection{Multi-Branch Architectures}
Multi-branch architecture is a common approach to combine features from different resolutions in visual perception, including image classification~\cite{chen2021crossvit}, object detection~\cite{HRnetv2, liang2021cbnetv2, vit_adapter, vit-comer}, semantic segmentation~\cite{yuan2021hrformer, gu2021hrvit} and multimodal understanding~\cite{llava-hr, hong2023cogagent}.
ViT-Adapter~\cite{vit_adapter} and ViT-CoMer~\cite{vit-comer} use a CNN-based branch to insert multi-scale information into a pretrained ViT branch for dense prediction tasks.
CBNetV2~\cite{liang2021cbnetv2} leverages an assistant backbone and a leading backbone in a Dense Higher Level Composition manner. These methods adopt asymmetric branch structures, leading to higher design complexity when extending to three or more branches.
CrossViT~\cite{chen2021crossvit} uses a two-branch structure with different patch sizes to obtain inputs of various scales, and leverages models of multiple sizes to balance the computational load. 
HRNet series~\cite{HRnetv2, yuan2021hrformer, gu2021hrvit} adopt a four-branch architecture for extracting high-resolution features, where the number of branches gradually increases as the layers deepen. 
However, they do not employ a parameter-inverted design and cannot utilize existing pretrained models.
In contrast, we propose a general visual perception architecture that utilizes pretrained models with different parameter scales to build efficient and powerful image pyramids.

\subsection{Multimodal Large Language Models}

With the rapid development of large language models (LLM)~\cite{VLM:GPT-4, VLM:Claude3, cai2024internlm2}, multimodal large language models (MLLM) have garnered significant research interest~\cite{VLM:GPT-4v, VLM:Gemini, internvl2.5}, which focus on utilizing pretrained LLMs for multimodal understanding and generation~\cite{emu3, synergen, janus, mm_interleaved, vlgpt}. Most existing MLLMs adopt a modular structure, \ie using a vision encoder to extract visual features from input images to be fed into the LLM~\cite{VLM:LLaVA-1.5, VLM:InternVL-1.5, bai2023qwenvl}. 
Some existing approaches attempt to equip MLLMs with high-resolution visual inputs for stronger multimodal understanding ability.
LLaVA-UHD~\cite{llava_uhd, llava_uhd_v2} employs an image modularization strategy to convert high-resolution images into condensed image tokens.
LLaVA-NeXT~\cite{VLM:LLaVA-NeXT} proposes an AnyRes strategy to use dynamic high resolution image inputs.
LLaVA-HR~\cite{llava-hr} uses a two-branch image pyramids with a specially designed MR-Adapter module for feature fusion.
Mini-Gemini~\cite{VLM:MiniGemini} adopts a similar two-branch design that injects the high-resolution feature maps into low-resolution ones with patch info mining.
MG-LLaVA~\cite{mg_llava} incorporates object-level features into MLLM through a  Multi-Granularity Vision Flow module to effectively process visual inputs of multiple granularities.
However, these methods use vision encoders with shared weights or same parameter scale to process large scale and small scale inputs, instead of the more efficient parameter-inverted design used in PIIP-LLaVA.
PIIP-LLaVA also employs cross-branch interactions to combine the multi-scale features for better representation learning.

\subsection{Redundancy Reduction for Visual Models.}
There have been extensive works focus on reducing computational redundancy to accelerate visual models, especially for ViTs. Some work attempts to reduce the number of visual tokens leveraging the sparsity of images to accelerate model inference. For instance, Dynamic ViT~\cite{rao2021dynamicvit} and AdaViT~\cite{meng2022adavit} predict and prune less informative tokens with lightweight prediction modules. EViT~\cite{liang2022not} and Evo-ViT~\cite{xu2022evo} identify less informative tokens and adopt accelerated processing strategies by computing attention scores for each token from the class token. 
Some studies refine the model structure for efficient computation, \eg attention mechanisms~\cite{wang2020linformer,gu2021hrvit,cai2023efficientvit}, or adopt a hierarchical architecture that gradually reduce the spatial resolution as the layers deepen~\cite{liu2021swin, pvt}.
Orthogonal to these studies, we propose to employ a parameter-inverted paradigm to avoid the computational cost of processing high-resolution images with large models.

\begin{figure*}[t]
    \centering
    \vspace{3mm}
    \includegraphics[width=\linewidth]{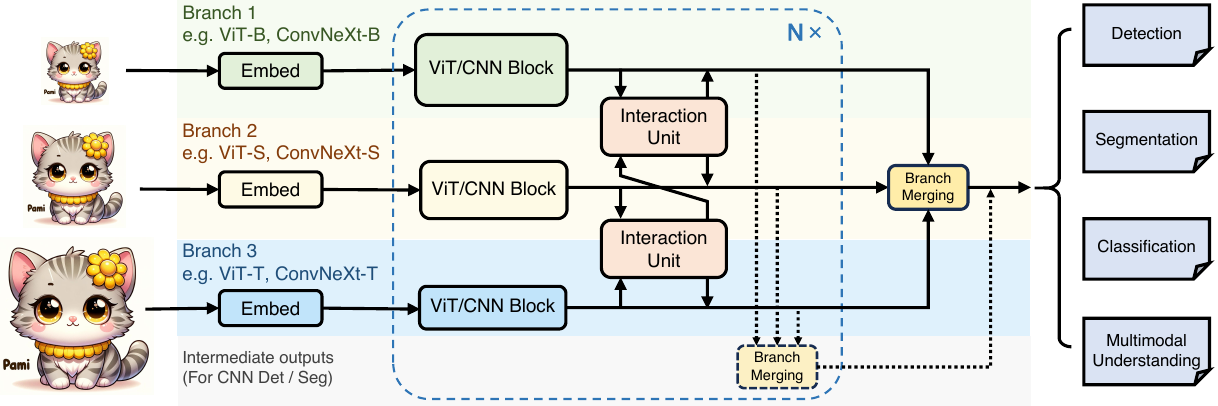}
        \vspace{1mm}
        \caption{\textbf{Overall architecture of PIIP.} We use multi-resolution branches to process images of different resolutions, where larger images are handled by smaller models. Each branch leverages pretrained ViTs or CNNs. Interaction units build connections between adjacent branches. Branch merging is inserted after all the blocks or within certain intermediate blocks to combine the features of all branches. 
        }
    \label{fig:architecture}
\end{figure*}

\begin{figure}[t]
  \centering
  \vspace{3mm}
  \includegraphics[width=\linewidth]{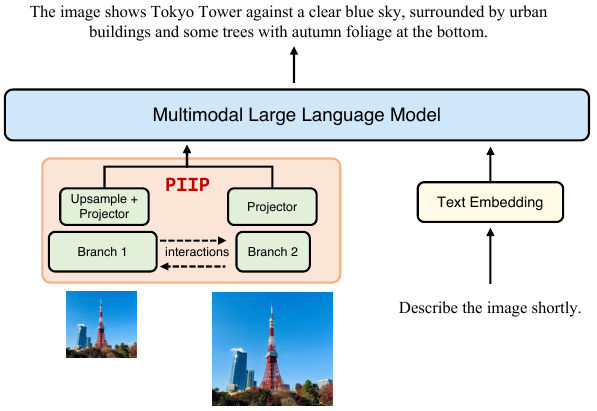}
   \caption{\textbf{Illustration of PIIP-LLaVA for multimodal understanding.} 
   We use one projector after each branch to align the visual features with the language embedding space of the LLM, and combine the features to obtain the visual features.
   }
   \label{fig:multimodal}
\end{figure}

\section{Methodology}
\label{sec:method}

To construct efficient and effective image pyramid networks, we employ a multi-branch structure in a parameter-inverted manner to handle images of different resolutions with models of different sizes. 
As shown in Figure~\ref{fig:architecture}, our proposed architecture comprises three components: multi-resolution branches, cross-branch interactions, and branch merging. Each branch uses an off-the-shelf pretrained ViT~\cite{vit} or CNN~\cite{convnext} model to process images of different resolutions, where larger resolutions are processed by branches with fewer parameters. Cross-branch interactions are inserted every few blocks to fuse features across different feature scales. Branch merging is inserted at the end or within intermediate blocks to combine the outputs from all branches. We use the existing pretrained ViTs~\cite{deit, deit3, augreg, VLP:CLIP} or CNNs~\cite{convnext} to initialize the branches, and initialize the interactions and branch merging from scratch.

\subsection{Multi-Resolution Branches}
\label{sec:method_branches}
The multi-resolution branches serve to extract visual representations from different image scales and semantic levels.
We first resize the input image to different resolutions through bilinear interpolation, which are then fed into corresponding branches of different scales. Each branch starts with an embedding module, which are patch embedding and position encoding for ViT~\cite{vit} or patchify stem for ConvNeXt~\cite{convnext}. The number of branches can be 2, 3 or 4.
All the branches have the same number of blocks $N$, where each block contains one or multiple ViT or CNN layers. Blocks from different branches usually have different feature dimensions due to the pretrained models, \eg ViT-T/ConvNeXt-T, ViT-S/ConvNeXt-S and ViT-B/ConvNeXt-B.
Branches with larger image sizes have a smaller number of parameters. For clarity, we refer to the branch with the largest number of parameters (with the smallest image size) as Branch 1, the second largest as Branch 2, and so on. The output of the $i$-th block of Branch $j$ is denoted as $\mathcal{F}^{i}_{j} \in \mathbb{R}^{H_jW_j / P_j^2 \times D_j}$, where $H_j$, $W_j$, $P_j$, $D_j$ are the image height, image width, patch size, and feature dimension of Branch $j$, respectively.

Typically, all branches are homogeneous, \eg ViT-T, ViT-S and ViT-B, or ConvNeXt-T, ConvNeXt-S and ConvNeXt-B.
Nevertheless, we also extend PIIP to heterogeneous structure, \ie some branches are ViTs while others are CNNs. Such structure has specific applications. CNN is capable of extracting local features from high-resolution inputs due to its locality and translation invariance, aligning with our intuition that high-resolution branch should focus on smaller receptive fields with detailed information. ViT has stronger global modeling ability when the context length is small, \ie low-resolution inputs, and can be employed to extract rich semantics. This also leverages the powerful semantic representation of ViTs with large-scale pretraining, \eg CLIP~\cite{VLP:CLIP} pretrained on the massive image-text paired data.

\subsection{Cross-Branch Interactions}
\label{sec:method_interaction}

\begin{figure*}[t]
    \centering
    \includegraphics[width=0.75\linewidth]{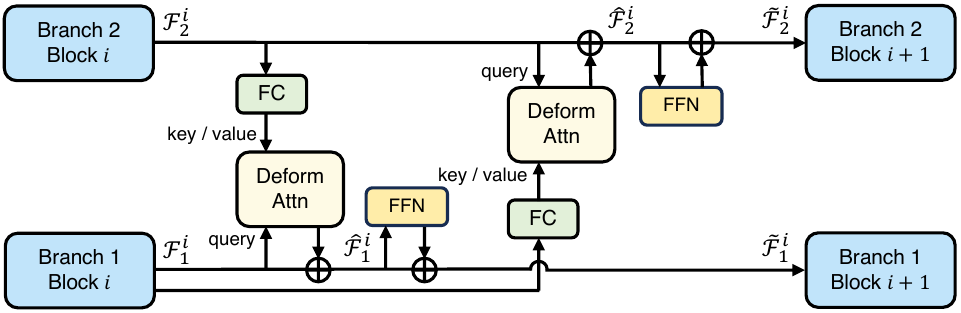}
    \caption{\textbf{Detailed structure of the interaction unit.} It consists of two deformable attentions with fully-connect layers and feed-forward networks.}
    \label{fig:interaction}
\end{figure*}

\begin{figure*}[t]
    \centering
    \includegraphics[width=\linewidth]{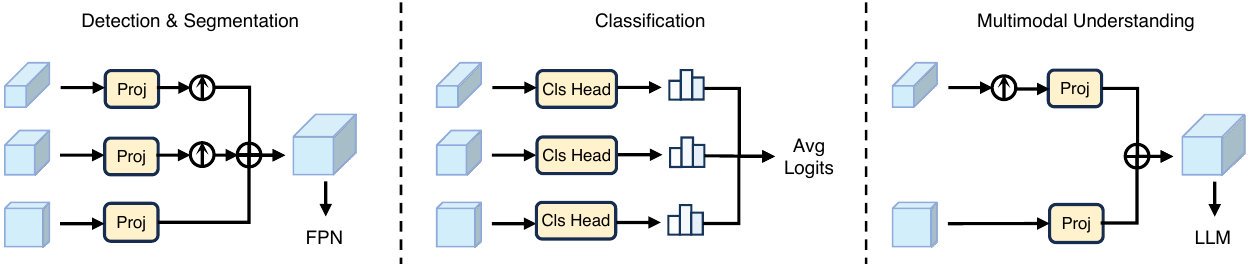}
    \caption{\textbf{Detailed design of branch merging in different tasks.} For detection, segmentation and multimodal understanding, output features from all branches are fused together with projection and upsampling, and fed into the subsequent FPN or LLM. For classification, we employ the original classification heads to compute logits, and average them as the final prediction.}
    \label{fig:merging}
\end{figure*}

Branches with different resolutions focus on different spatial scales and semantic levels. To complement the features of various scales, we propose the cross-branch interactions. Each cross-branch interaction consists of several interaction \textit{units}, where each unit builds connections between outputs from two feature-scale adjacent branches. The structure of an interaction unit is shown in Figure~\ref{fig:interaction}. 

Specifically, for the outputs of the $i$-th block of Branch 1 and 2, denoted as $\mathcal{F}^{i}_{\rm 1} \in \mathbb{R}^{H_1W_1 / P_1^2 \times D_1}$ and $\mathcal{F}^{i}_{\rm 2} \in \mathbb{R}^{H_2W_2 / P_2^2 \times D_2}$, we perform two deformable cross-attention~\cite{deformable_detr} between the two hidden features, denoted as ${\rm Attn}(\cdot)$. Each cross attention is preceded by a linear layer ${\rm FC}(\cdot)$ to project the feature dimension of key and value into that of the query, \ie from $D_1$ to $D_2$ or vice versa. A feed-forward network ${\rm FFN}(\cdot)$ is added after each cross attention to provide channel-wise feature fusion. The hidden dimension ratio of FFN is set to $0.25$ to save computational costs.

For the first cross-attention in the interaction unit, the interaction process can be formulated as:
\begin{align}
	\mathcal{\hat{F}}^{i}_{\rm 1}&=\mathcal{F}^{i}_{\rm 1} + \gamma^{i}_{\rm 1} {\rm Attn}({\rm norm}(\mathcal{F}^{i}_{\rm 1}), {\rm norm}({\rm FC}(\mathcal{F}^{i}_{\rm 2}))), \\
    \mathcal{\tilde{F}}^{i}_{\rm 1}&=\mathcal{\hat{F}}^{i}_{\rm 1} + \tau^{i}_{\rm 1} {\rm FFN}({\rm norm}(\mathcal{\hat{F}}^{i}_{\rm 1})),
\end{align}
where ${\rm norm}(\cdot)$ is LayerNorm~\cite{layernorm}, $\tau^{i}_{\rm 1}$ and $\gamma^{i}_{\rm 1}$ are learnable parameters, and $\mathcal{\tilde{F}}^{i}_{\rm 1}$ is the interaction output. 
$\tau^{i}_{\rm 1}$ and $\gamma^{i}_{\rm 1}$ are initialized with $\mathbf{0}$ to ensure that the feature extraction of the original blocks (\ie distribution of $\mathcal{F}^{i}_{\rm 1}$) will not be modified drastically during the interactions, thereby better leveraging the pretrained weights.

Similarly, the second cross-attention is performed by switching the query and key/value to obtain $\mathcal{\tilde{F}}^{i}_{\rm 2}$.
The outputs $\mathcal{\tilde{F}}^{i}_{\rm 1}$ and $\mathcal{\tilde{F}}^{i}_{\rm 2}$ are used for subsequent feature extractions.
We only construct interaction units between each pair of feature-scale adjacent branches, such as Branch 1 \& Branch 2 and Branch 2 \& Branch 3 for a three-branch network.
\begin{table*}[]
\centering
\caption{\textbf{Comparison with baseline on COCO val2017.} We report the number of parameters and FLOPs of the backbone. \underline{Underline} indicates FLOPs or metrics on par with the baseline. $\rm AP^b$ and $\rm AP^m$ represent box AP and mask AP, respectively.}
\renewcommand\arraystretch{1.2}
\resizebox{0.95\linewidth}{!}{
    \begin{tabular}{c|ccc|cccccc}
    \toprule
    \multirow{2}{*}{\textbf{Model}} & \multirow{2}{*}{\textbf{Resolution}} & \multirow{2}{*}{\textbf{\#Param}} & \multirow{2}{*}{\textbf{\#FLOPs}} & \multicolumn{6}{c}{\textbf{Mask R-CNN 1$\times$ schedule}} \\ 
    & & & & $\rm AP^b$ & $\rm AP^b_{50}$ & $\rm AP^b_{75}$ & $\rm AP^m$ & $\rm AP^m_{50}$ & $\rm AP^m_{75}$ \\
    \midrule
    ViTDet-B \cite{vitdet} & 1024 & 90M & 463G & 43.8 & 67.6 & 47.7 & 39.9 & 63.6 & 42.2  \\
    \rowcolor{gray!10} & 1120/896/448 & 146M & 243G & \underline{43.9} & 65.7 & 47.5 & \color{gray!70}{38.6} & \color{gray!70}{61.8} & \color{gray!70}{40.6}  \\
    \rowcolor{gray!10} PIIP-TSB (ours) & 1568/896/448 & 147M & 287G & \color{gray!70}{45.0} & \color{gray!70}{67.0} & \color{gray!70}{48.7} & \underline{40.2} & 63.8 & 42.6  \\
    \rowcolor{gray!10} & 1568/1120/672 & 149M & \underline{453G} & \textbf{46.6} & 68.4 & 51.1 & \textbf{41.4} & 65.2 & 44.3\vspace{-0.6mm}\\ 
    \midrule
    ViTDet-L \cite{vitdet} & 1024 & 308M & 1542G & 46.8 & 70.8 & 51.4 & 42.5 & 67.3 & 45.3 \\
    \rowcolor{gray!10} & 1120/672/448 & 493M & 727G & \underline{46.7} & 69.0 & 50.6 & \color{gray!70}{40.8} & \color{gray!70}{65.2} & \color{gray!70}{42.8} \\
    \rowcolor{gray!10} PIIP-SBL (ours) & 1344/896/448 & 495M & 1002G & \color{gray!70}{48.2} & \color{gray!70}{71.0} & \color{gray!70}{52.8} & \underline{42.5} & 67.3 & 45.4 \\
    \rowcolor{gray!10} & 1568/896/672 & 497M & \underline{1464G} & 49.4 & 71.9 & 53.9 & 43.7 & 68.4 & 46.6 \\
    \rowcolor{gray!10} \vspace{-2.5mm} & & & & & & & & & \\
    \rowcolor{gray!10} & 1344/896/672/448 & 506M & 755G & \underline{46.9} & 69.9 & 50.6 & \color{gray!70}{41.6} & \color{gray!70}{65.9} & \color{gray!70}{44.1} \\
    \rowcolor{gray!10} PIIP-TSBL (ours) & 1568/1120/672/448 & 507M & 861G & \color{gray!70}{48.2} & \color{gray!70}{70.5} & \color{gray!70}{52.7} & \underline{42.8} & 66.9 & 45.6 \\
    \rowcolor{gray!10} & 1792/1568/1120/448 & 512M & \underline{1535G} & \textbf{49.6} & 72.4 & 54.2 & \textbf{44.2} & 69.2 & 47.5\vspace{-0.6mm}\\
    \bottomrule
    \end{tabular}
}
\label{table:baseline}
\end{table*}
\subsection{Branch Merging}
\label{sec:method_merging}

The final feature maps of all branches $\mathcal{\tilde{F}}^{N}_{j}$ have different spatial shapes and feature dimensions, where spatially larger feature maps have fewer feature dimensions. 
Since a single feature map fails to provide multi-scale semantic features, we employ branch merging to merge the outputs of all branches into a single feature map. In most scenarios, branch merging is added only after all blocks to form the final output. However, for CNN-based detection and segmentation where hierarchical features are required, branch merging is also inserted within intermediate blocks to obtain intermediate outputs, as depicted in the bottom of Figure~\ref{fig:architecture}. This resembles the classical design of CNN-based detectors~\cite{fpn, internimage} where features from middle stages are fed into the feature pyramid network (FPN).

The detailed structure of branch merging is illustrated in Figure~\ref{fig:merging}. We elaborate the design in detection and segmentation as an example. All branch outputs except for Branch 1 are first projected to the feature dimension of Branch 1 (the largest feature dimension) with ${\rm Proj}(\cdot)$, which comprises two convolutional layers and GroupNorm~\cite{group_norm}.
Then, all branch outputs are upsampled by bilinear interpolation ${\rm Upsample}(\cdot)$ into the feature map size of the last branch (the largest feature map size). 
Finally, these outputs, with the same spatial shape and feature dimension, are aggregated with learnable scalar weights $w_j$ to form the final output. This process can be formulated as:

\begin{align}
	\mathcal{\tilde{F}}^{\rm out}_{j} &= {\rm Upsample} ( {\rm Proj}(\mathcal{\tilde{F}}^{N}_{j}) ), \\
	\mathcal{F}^{\rm out} &= \sum_{j=1}^M w_j \mathcal{\tilde{F}}^{\rm out}_{j},
\end{align}

where $M$ is the number of branches. 
$\mathcal{F}^{\rm out}$ is the output feature map, which has the largest feature resolution and also the largest feature dimension across all branches.
It subsequently serves as input to FPN~\cite{fpn} as in common detection and segmentation models. 

For image classification, we do not use a branch merging module, but append the original classification heads of the pretrained models after each branch. The final predicted score of each class is computed as the average of the output logits of all branches. We observe that using pretrained heads can speed up convergence compared to using a randomly initialized head after branch merging.

For multimodal understanding, common modular MLLM architectures~\cite{llava, internvl2.5, bai2023qwenvl} employ a projector between the vision encoder and language model~\cite{llava, VLM:LLaVA-1.5} to align the visual features with the LLM embedding space. 
In our MLLM called PIIP-LLaVA, we add a separate projector for each branch inside the branch merging module.
The overall structure of PIIP-LLaVA is shown in Figure~\ref{fig:multimodal}.

\begin{figure*}[!tb]
    \centering
    \subfigure[Object detection]{
        \begin{minipage}[t]{0.4\linewidth}
            \centering
            \includegraphics[width=1.0\linewidth]{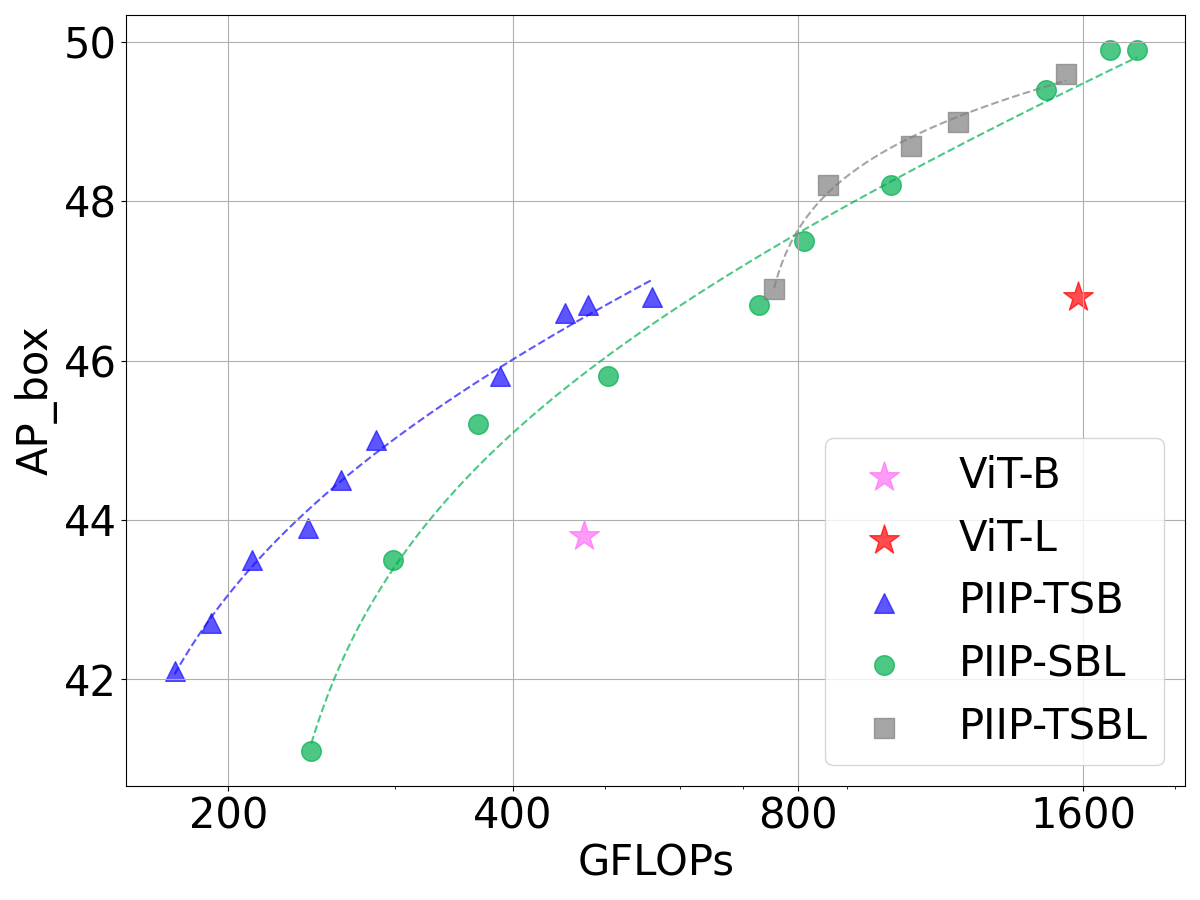}
        \end{minipage}
        \label{fig:baseline_box}
    }
    \subfigure[Instance segmentation]{
        \begin{minipage}[t]{0.4\linewidth}
            \centering
            \includegraphics[width=1.0\linewidth]{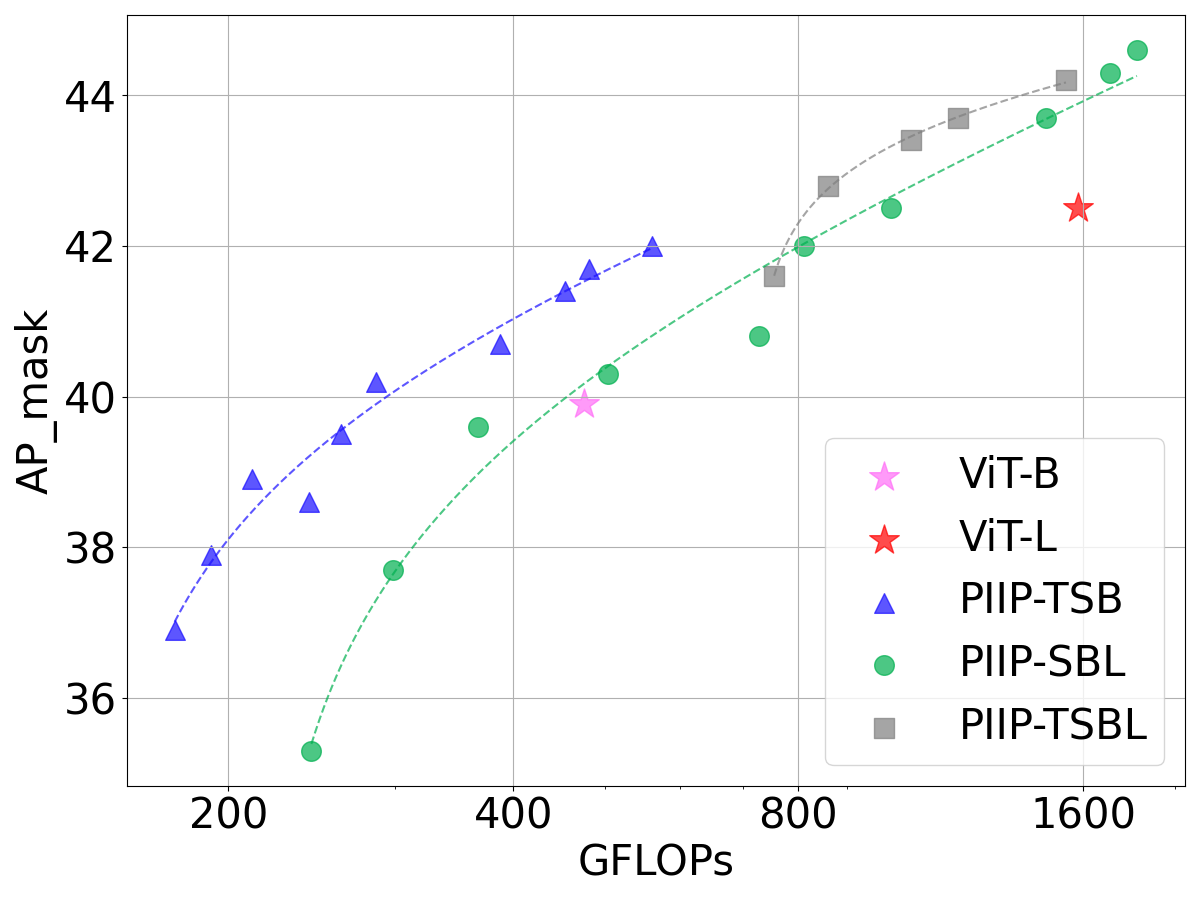}
        \end{minipage}
        \label{fig:baseline_mask}
    }
    \caption{\textbf{Performance of different PIIP variants by adjusting input resolutions on object detection and instance segmentation.}
    }
    \label{fig:baseline}
\end{figure*}

\begin{table*}[]
\centering
\caption{\textbf{Performance of CNN-based and heterogeneous models on COCO val2017.}}
\renewcommand\arraystretch{1.4}
\resizebox{0.9\linewidth}{!}{
    \begin{tabular}{c|ccc|cc|cc}
    \toprule
    \multirow{2}{*}{\textbf{Model}} & \multicolumn{3}{c}{\textbf{Branches}} & \multirow{2}{*}{\textbf{Resolution}} & \multirow{2}{*}{\textbf{\#FLOPs}} & \multicolumn{2}{c}{\textbf{Mask R-CNN 1$\times$ schedule}} \\ 
    & Tiny & Small & Base & & & $\rm AP^b$ & $\rm AP^m$ \\
    \midrule
    ViTDet-B \cite{vitdet} & $-$ & $-$ & ViT & 1024 & 463G & 43.8 & 39.9 \\
    ConvNeXt-B \cite{convnext} & $-$ & $-$ & ConvNeXt & 1024 & 321G & 42.4 & 38.7 \\
    \rowcolor{gray!10} & ViT & ViT & ViT & 1024/672/448 & 225G & 44.3 & 39.9  \\
    \rowcolor{gray!10} & ConvNeXt & ConvNeXt & ConvNeXt & 1024/672/448 & 326G & 46.4 & 41.7  \\
    \rowcolor{gray!10} & ConvNext & ViT & ViT & 1024/672/448 & 373G & 47.1 & 42.4  \\
    \rowcolor{gray!10} & ConvNext & ConvNext & ViT & 1024/672/448 & 431G & 46.8 & 42.2  \\ 
    \rowcolor{gray!10} & ConvNext & ViT & ConvNext & 1024/672/448 & 297G & 46.7 & 42.0  \\
    \rowcolor{gray!10} & ViT & ConvNext & ViT & 1024/672/448 & 291G & 45.4 & 40.9  \\
    \rowcolor{gray!10} & ViT & ConvNext & ConvNext & 1024/672/448 & 231G & 45.2 & 40.7  \\
    \rowcolor{gray!10} \multirow{-8}{*}{PIIP-TSB (ours)} & ViT & ViT & ConvNext & 1024/672/448 & 193G & 44.8 & 40.3\vspace{-0.6mm} \\
    \bottomrule
    \end{tabular}
}
\label{table:cnn}
\end{table*}

\begin{table*}[t]
\caption{\textbf{Object detection and instance segmentation performance on COCO val2017.} `MS' means using AutoAugment~\cite{autoaugment} for multi-scale training. Large-size models use ViT weights trained on ImageNet-21K. 
Some results are from~\cite{vit_adapter}. 
}
\begin{minipage}[t]{0.49\linewidth}
\centering
\renewcommand\arraystretch{1.2}
\resizebox{\columnwidth}{!}{
    \begin{tabular}{c|cccccc}
    \toprule
    \textbf{Method} & \textbf{$\rm AP^b$} & \textbf{$\rm AP^b_{50}$} & \textbf{$\rm AP^b_{75}$} & \textbf{$\rm AP^m$} & \textbf{$\rm AP^m_{50}$} & \textbf{$\rm AP^m_{75}$} \\
    \midrule
    & \multicolumn{6}{c}{\textbf{Mask R-CNN 1$\times$ schedule}} \\
    PVTv2-B5~\cite{pvt} & 47.4 & 68.6 & 51.9 & 42.5 & 65.7 & 46.0 \\
    ViT-B~\cite{li2021benchmarking} & 42.9 & 65.7 & 46.8 & 39.4 & 62.6 & 42.0 \\
    ViTDet-B \cite{vitdet} & 43.2 & 65.8 & 46.9 & 39.2 & 62.7 & 41.4 \\
    Swin-B~\cite{liu2021swin} & 46.9 & - & - & 42.3 & - & - \\
    ViT-Adapter-B~\cite{vit_adapter} & 47.0 & 68.2 & 51.4 & 41.8 & 65.1 & 44.9 \\
    \rowcolor{gray!10} PIIP-TSB (ours) & 47.9 & 70.2 & 52.5 & 42.6 & 67.2 & 45.5\vspace{-0.6mm}  \\
    \midrule
    ViT-L \cite{li2021benchmarking} & 45.7 & 68.9 & 49.4 & 41.5 & 65.6 & 44.6 \\
    ViTDet-L \cite{vitdet} & 46.2 & 69.2 & 50.3 & 41.4 & 65.8 & 44.1 \\
    ViT-Adapter-L \cite{vit_adapter} & 48.7 & 70.1 & 53.2 & 43.3 & 67.0 & 46.9 \\
    \rowcolor{gray!10} PIIP-SBL (ours) & 49.9 & 72.8 & 54.7 & 44.6 & 69.3 & 47.9\vspace{-0.6mm} \\
    \midrule
    & \multicolumn{6}{c}{\textbf{DINO + MS schedule}} \\
    \rowcolor{gray!10} PIIP-SBL-3$\times$ (ours) & 57.9 & 76.9 & 63.3 & - & - & - \\
    \rowcolor{gray!10} PIIP-H6B-1$\times$ (ours) & 60.0 & 79.0 & 65.4 & - & - & -\vspace{-0.6mm} \\
    \bottomrule
    \end{tabular}
}
\end{minipage}
\begin{minipage}[t]{0.49\linewidth}
\centering
\renewcommand\arraystretch{1.2}
\resizebox{\columnwidth}{!}{
    \begin{tabular}{c|cccccc}
    \toprule
    \textbf{Method} & \textbf{$\rm AP^b$} & \textbf{$\rm AP^b_{50}$} & \textbf{$\rm AP^b_{75}$} & \textbf{$\rm AP^m$} & \textbf{$\rm AP^m_{50}$} & \textbf{$\rm AP^m_{75}$} \\
    \midrule
    & \multicolumn{6}{c}{\textbf{Cascade R-CNN 1$\times$ schedule}} \\
    Swin-L~\cite{liu2021swin} & 51.8 & 71.0 & 56.2 & 44.9 & 68.4 & 48.9 \\
    ConvNeXt-L~\cite{convnext} & 53.5 & 72.8 & 58.3 & 46.4 & 70.2 & 50.2 \\
    \rowcolor{gray!10} PIIP-SBL (ours) & 53.6 & 73.3 & 57.9 & 46.3 & 70.3 & 50.0\vspace{-0.6mm} \\
    \midrule
    & \multicolumn{6}{c}{\textbf{Cascade R-CNN 3$\times$ + MS schedule}} \\
    Swin-B~\cite{liu2021swin} & 51.9 & 70.9 & 57.0 & - & - & - \\
    Shuffle-B~\cite{huang2021shuffle} & 52.2 & 71.3 & 57.0 & - & - & - \\
    ViT-B~\cite{li2021benchmarking} & 50.1 & 69.3 & 54.3 & - & - & - \\
    ViT-Adapter-B~\cite{vit_adapter} & 52.1 & 70.6 & 56.5 & - & - & - \\
    \rowcolor{gray!10} PIIP-TSB (ours) & 53.1 & 72.3 & 57.4 & 46.5 & 70.1 & 51.1\vspace{-0.6mm} \\
    \midrule
    Swin-L~\cite{liu2021swin} & 53.9 & 72.4 & 58.8 & 46.7 & 70.1 & 50.8 \\
    RepLKNet-31L~\cite{ding2022scaling} & 53.9 & 72.5 & 58.6 & 46.5 & 70.0 & 50.6 \\
    ConvNeXt-L~\cite{convnext} & 54.8 & 73.8 & 59.8 & 47.6 & 71.3 & 51.7 \\
    \rowcolor{gray!10} PIIP-SBL (ours) & 54.5 & 73.8 & 59.1 & 47.7 & 71.6 & 52.1\vspace{-0.6mm} \\
    \bottomrule
    \end{tabular}
}
\end{minipage}
\label{table:detection}
\end{table*}

\section{Experiments}
\label{sec:experiments}

\vspace{1mm}
\subsection{Implementation Details}
\label{sec:exp_implement}
For comparison with base-size models, we use pretrained ViT-T/S/B as the branches to construct three-branch PIIP network PIIP-TSB. Similarly, ViT-S/B/L are used to construct PIIP-SBL to match the computation of large-size models. We also construct four-branch PIIP-TSBL model. For CNN-based and heterogeneous models, we employ ConvNeXt~\cite{convnext} as the branches. Unless specified, PIIP models are ViT-based in perception tasks, \ie detection, segmentation, and classification.

We set the number of interactions (each with 2 interaction units as shown in Figure~\ref{fig:architecture}) $N$ to 12, \ie every layer for ViT-T/S/B, every two layers for ViT-L, or every three layers for ConvNeXt-S/B/L. We construct multiple variants of two-branch, three-branch and four-branch models with different resolution configurations. For combinations with an inconsistent number of layers, we adopt a larger learning rate decay for the backbone with fewer layers. For example, for ViT-S/B (12 layers) and ViT-L (24 layers), the learning rate decay for ViT-S/B is set to be twice that of ViT-L (24/12=2).

For object detection and segmentation, we use ViT-S/B/L pretrained on ImageNet~\cite{imagenet} from DeiT III~\cite{deit3}, ViT-T from DeiT~\cite{deit}. ViT-H from MAE~\cite{mae} and InternViT-6B~\cite{internvl} are used for 6B-scale experiments. For all PIIP-SBL models, we use the ImageNet-21K 384-resolution pretrained weights to compare with previous approaches. We adopt AdamW~\cite{adamw} optimizer with layer-wise learning rate decay~\cite{bao2021beit} to train the model on 8 NVIDIA A800 GPUs. 
For image classification, in base-size experiments we use pretrained ViT-T/S/B weights from DeiT~\cite{deit}. In large-size experiments, since DeiT does not provide ViT-L models, we use ImageNet-21K pretrained ViT-S/B/L weights from~\cite{augreg}.
For multimodal understanding, we CLIP-B/L from~\cite{VLP:CLIP}
and ConvNeXt-B/L pretrained on Laion Aesthetic~\cite{Datasets:Laion-5b}.

We use the FLOPs calculation script from MMDetection~\cite{mmdetection}, with our modifications to accurately calculate FLOPs of self-attention and deformable attention modules. The script is released along with the training code. We have also manually verified the calculations using formulas, and the results are consistent with those produced by the script.
\begin{table*}[!tb]
\small
\centering
\caption{\textbf{Experiments on the large-scale vision foundation model InternViT-6B.}}
\renewcommand\arraystretch{1.2}
\resizebox{0.9\linewidth}{!}{
    \begin{tabular}{cc|cccc|ccc}
    \toprule
    \multirow{2}{*}{\textbf{Model}} & \multirow{2}{*}{\textbf{\#Param}} & \multicolumn{4}{c|}{\textbf{Mask R-CNN 1$\times$ schedule}} & \multicolumn{3}{c}{\textbf{UperNet 160k}} \\ 
    & & \#FLOPs & Resolution & $\rm AP^b$ & $\rm AP^m$ & Crop Size & \#FLOPs & mIoU \\
    \midrule
    InternViT-6B \cite{internvl}& 5919M & 24418G & 1024 & 53.8 & 48.1 & 512 & 6105G & 58.36  \\
    \rowcolor{gray!10} & 7269M & 5643G & 1280/1024/256 & 53.5 & 47.5  & 640/512/192 & 1903G & 57.82 \\
    \rowcolor{gray!10} PIIP-LH6B (ours) & 7271M & 10368G & 1280/1024/512 & 54.4 & 47.8 & 640/512/256 & 2592G & 58.42 \\
    \rowcolor{gray!10} & 7273M & 13911G & 1280/1024/640 & \textbf{55.7} & \textbf{49.0} & 640/512/384 & 4560G & \textbf{59.65}\vspace{-0.6mm} \\
    \bottomrule
    \end{tabular}
}
\label{table:internvit_6b}
\end{table*}
\noindent
\begin{table*}[t]
\begin{minipage}[t]{0.45\textwidth}
    \small
    \centering
    \caption{\textbf{Experiments of initializing with different pre-trained weights on COCO val2017} with PIIP-SBL 1568/1120/672.}
    \vspace{-1mm}
    \renewcommand\arraystretch{1.3}
    \resizebox{0.94\textwidth}{!}{
        \begin{tabular}{cc|cc}
        \toprule
        \textbf{ViT-S} & \textbf{ViT-B / ViT-L} & $\rm AP^b$ & $\rm AP^m$ \\
        \midrule
        AugReg~\cite{augreg} & AugReg~\cite{augreg} & 48.3 & 42.6  \\
        DeiT III~\cite{deit3} & Uni-Perceiver~\cite{uni_perceiver} & 48.8 & 42.9  \\
        DeiT III~\cite{deit3} & MAE~\cite{mae} & 49.1 & 43.0  \\
        DeiT III~\cite{deit3} & DeiT III~\cite{deit3} & 50.0 & 44.4  \\
        DeiT III~\cite{deit3} & DINOv2~\cite{dinov2} & 51.0 & 44.7  \\
        DeiT III~\cite{deit3} & BEiTv2~\cite{beitv2} & 51.8 & 45.4  \\
        \bottomrule
        \end{tabular}
    }
    \vspace{3mm}
    
    \label{table:det_pretrain}
    \centering
    \caption{\textbf{Comparison with baseline on ADE20K using UperNet.}}
    \vspace{-1mm}
    \renewcommand\arraystretch{1.3}
    \resizebox{0.9\textwidth}{!}{
        
        \begin{tabular}{c|c|cc}
        \toprule
        \textbf{Method} & \textbf{Crop Size} & \textbf{\#FLOPS} & \textbf{mIoU} \\
        \midrule
        ViT-B & 640 & 159G & 51.0 \\
        \rowcolor{gray!10} PIIP-TSB (ours) & 896/448/336 & 118G & 51.6\vspace{-0.6mm} \\
        \midrule
        ViT-L & 640 & 545G & 53.6 \\
        \rowcolor{gray!10} PIIP-SBL (ours) & 1120/448/336 & 456G & 54.3\vspace{-0.6mm}  \\
        \bottomrule
        \end{tabular}
    }
    \label{table:segmentation_baseline}
\end{minipage}
\quad \quad
\begin{minipage}[t]{0.5\textwidth}
    \centering
    \caption{\textbf{Semantic segmentation performance on ADE20K using UperNet.}}
    \renewcommand\arraystretch{1.5}
    \resizebox{0.79\textwidth}{!}{
        \begin{tabular}{c|c|c}
        \toprule
        \textbf{Method} & \textbf{Crop Size} & \textbf{mIoU} \\
        \midrule
        Swin-B \cite{liu2021swin} & 512 & 48.1 \\
        ConvNeXt-B \cite{convnext} & 512 & 49.1 \\
        RepLKNet-31B \cite{ding2022scaling} & 512 & 49.9 \\
        SLaK-B \cite{liu2022more} & 512 & 50.2 \\
        InternImage-B \cite{internimage} & 512 & 50.2 \\
        \rowcolor{gray!10} PIIP-TSB (ours) & 896/448/336 & \textbf{51.6}\vspace{-0.6mm} \\
        \midrule
        Swin-L \cite{liu2021swin} & 640 & 52.1 \\
        RepLKNet-31L \cite{ding2022scaling} & 640 & 52.4 \\
        ConvNeXt-L \cite{convnext} & 640 & 53.2 \\
        ConvNeXt-XL \cite{convnext} & 640 & 53.6 \\
        InternImage-L \cite{internimage} & 640 & 53.9 \\
        \rowcolor{gray!10} PIIP-SBL (ours) & 1120/448/336 & \textbf{54.3}\vspace{-0.6mm} \\
        \bottomrule
        \end{tabular}
    }
    \label{table:segmentation}
\end{minipage}
\end{table*}
\subsection{Object Detection and Instance Segmentation}
\label{sec:exp_det}

\textbf{Setting.} 
We conduct object detection and instance segmentation experiments with MMDetection~\cite{mmdetection} on MS COCO dataset~\cite{coco}.
Three detectors are used, including Mask R-CNN~\cite{mask_rcnn}, Cascade R-CNN~\cite{cai2018cascade} and DINO~\cite{zhang2022dino}. Following common practices~\cite{vit_adapter}, we adopt 1$\times$ (12 epochs) or 3$\times$ (36 epochs) training schedules and use window attention~\cite{vitdet} to save time and memory. The total batch size is 16, and the initial learning rate and weight decay are 1e-4 and 0.05.

\textbf{Comparison with baseline.}
To demonstrate the performance and computational advantages of PIIP, we validate the effectiveness of PIIP against two baseline models: ViTDet-B and ViTDet-L~\cite{vitdet}, as shown in Table~\ref{table:baseline}. While maintaining similar performance with ViTDet-B, our PIIP-TSB reduces the computational cost by 47.5\% (243G vs. 463G) and 38.0\% (287G vs. 463G) in object detection and instance segmentation tasks respectively. Similarly, compared with ViTDet-L, our PIIP-SBL reduces the computational cost by about 52.9\% (727G vs. 1542G) and 35.0\% (1002G vs. 1542G) in the two tasks. On the other hand, when keeping similar computational costs as the baseline, PIIP-TSB and PIIP-SBL improve the object detection performance by 2.8\% and 2.6\%, respectively, and instance segmentation by 1.5\% and 1.2\%, compared to ViTDet-B and ViTDet-L. 
To better understand the above conclusion, we depict the trend between the computational cost and performance of different PIIP variants by adjusting the input resolution, as shown in Figure~\ref{fig:baseline}. Furthermore, the performance curve for the four-branch structure demonstrates a slight improvement over the three-branch structure. 

\textbf{Results of CNN-based and heterogeneous models.} As given in Table~\ref{table:cnn}, we study the effectiveness of CNN-based and CNN-ViT hybrid models. We find that the three-branch ConvNeXt TSB model outperforms the ConvNeXt-B baseline with equivalent computation, demonstrating that PIIP can also effectively reduce computation and improve performance on CNN image pyramids. As for heterogenous structures, models with ConvNeXt for higher resolution and ViT for lower resolution achieve the best performance, consistent with our observation in Section~\ref{sec:exp_multimodal}.

\textbf{Results with base-size and large-size models.} As shown in Table~\ref{table:detection}, combined with Mask R-CNN, PIIP achieves higher performance than ViT-Adapter by a considerable margin, about 0.9\% and 1.2\% on $\rm AP^b$. With a more powerful detector Cascade R-CNN and stronger training schedule (3$\times$ + MS), PIIP-TSB and PIIP-SBL achieve competitive performance of 53.1\% and 54.5\% $\rm AP^b$, respectively. Finally, we achieve 60.0\% $\rm AP^b$ with the DINO~\cite{zhang2022dino} detector.

\textbf{Results with InternViT-6B.}
In Table~\ref{table:internvit_6b}, we further examine PIIP on an extremely large vision foundation model InternViT-6B~\cite{internvl}, and achieve 55.7\% $\rm AP^b$ using Mask R-CNN 1$\times$ training schedule. In addition, PIIP can save nearly 43\% of the computation and achieve better performance than the single-branch InternViT-6B by 1.9\% on $\rm AP^b$ and 0.9\% on $\rm AP^m$.

\textbf{Results with different pretraining methods.} To study the influence of different pretraining weights, we initialize PIIP-SBL with pretrained ViT S/B/L weights from AugReg~\cite{augreg}, Uni-Perceiver~\cite{uni_perceiver}, MAE~\cite{mae}, DeiT III~\cite{deit3}, DINOv2~\cite{dinov2} and BEiTv2~\cite{beitv2}. As shown in Table~\ref{table:det_pretrain}, the BEiTv2-initialized model achieves the best performance.

\subsection{Semantic Segmentation}
\label{sec:exp_seg}
\begin{table*}[]
\begin{minipage}[t]{0.63\linewidth}
\centering
\caption{\textbf{Image classification performance on ImageNet.} \underline{Underline} indicates FLOPs or metrics on par with the baseline.}
\renewcommand\arraystretch{1.2}
\resizebox{0.85\linewidth}{!}{
    \begin{tabular}{c|ccc}
    \toprule
    \textbf{Model} & \textbf{Resolution} & \textbf{\#FLOPs} & \textbf{Top-1 Acc} \\ 
    \midrule
    DeiT-B~\cite{deit} & 224 & 17.2G & 81.8  \\
    \rowcolor{gray!10} PIIP-TSB (ours) & 368/192/128 & \underline{17.4G} & 82.1\vspace{-0.6mm}  \\
    \midrule
    ViT-L~\cite{augreg} & 224 & 61.6G & 84.0 \\
    ViT-L~\cite{augreg}  (our impl.) & 224 & 61.6G & 85.2 \\
    \rowcolor{gray!10} PIIP-SBL (ours) & 320/160/96 & 39.0G & \underline{85.2}  \\
    \rowcolor{gray!10} PIIP-SBL (ours) & 384/192/128 & \underline{61.2G} & 85.9\vspace{-0.6mm} \\
    \bottomrule
    \end{tabular}
}
\label{table:cls_baseline}
\end{minipage}
\quad \quad
\begin{minipage}[t]{0.35\linewidth}
\centering
\caption{\textbf{Ablation on Branch Merging on COCO val2017 with PIIP-TSB 1568/896/672.}}
\renewcommand\arraystretch{1.02}
\resizebox{0.83\linewidth}{!}{
    \begin{tabular}{c|cc}
    \toprule
    \textbf{Out Branch} & \textbf{$\rm AP^b$} & \textbf{$\rm AP^m$} \\ 
    \midrule
    B & 43.1 & 37.0  \\
    S & 44.7 & 39.1 \\
    T & 45.6 & 40.6 \\
    B+S & 45.4 & 39.8  \\
    B+T & 46.3 & 41.1 \\
    S+T & 46.2 & 40.9 \\
    B+S+T & \textbf{46.6} & \textbf{41.4} \\
    \bottomrule
    \end{tabular}
}
\label{table:merge}
\end{minipage}
\end{table*}
\begin{table*}[]
\centering
\caption{\textbf{From-scratch pre-training settings and results on ImageNet-1K.}}
\renewcommand\arraystretch{1.2}

\subfigure[Configuration]{
    \resizebox{0.65\linewidth}{!}{ 
        \begin{tabular}{c|cccc|cc}
        \toprule
        \textbf{Module} & \textbf{\#Layers} & \textbf{Dim} & \textbf{\#Heads} & \textbf{Resolution} & \textbf{\#Param} & \textbf{\#FLOPs} \\
        \midrule
        Branch 1 & 12 & 640 & 8 & 128 & 59.6M & 3.8G \\
        Branch 2 & 12 & 320 & 4 & 256 & 15.1M & 4.3G \\
        Branch 3 & 12 & 160 & 2 & 512 & 4.0M & 4.9G \\
        Interactions & 12 & - & - & - & 21.2M & 5.1G \\
        Branch Merging & - & - & - & - & 0.3M & 0.2G \\
        \bottomrule
        \end{tabular}
    }
}

\subfigure[Performance]{
\renewcommand\arraystretch{1.3}
\resizebox{0.55\linewidth}{!}{
        \begin{tabular}{c|cccc}
        \toprule
        \textbf{Model} & \textbf{Resolution} & \textbf{\#Param} & \textbf{\#FLOPs} & \textbf{Top-1 Acc} \\
        \midrule
        ViT-B (our impl.) & 224 & 86M & 17.5G & 82.0  \\
        \rowcolor{gray!10} PIIP-B (ours) & 512/256/128 & 100M & 18.4G & 82.7\vspace{-0.6mm}  \\
        \bottomrule
        \end{tabular}
    }
}

\label{table:pretraining}
\end{table*}
\noindent
\begin{table*}[t]
    \small
    \centering
    \caption{\textbf{Comparison with multi-resolution baselines on multimodal benchmarks.} All models are trained with LLaVA-1.5~\cite{VLM:LLaVA-1.5} training data. We report \#FLOPs of the vision encoder. 
    }
    \vspace{-1mm}
    \renewcommand\arraystretch{1.55}
    \resizebox{\textwidth}{!}{
        
        \begin{tabular}{cc|ccc|cccccccc|c}
        \toprule
        \textbf{Method} & \textbf{Figure} & \textbf{Vision Encoder} & \textbf{Resolution} & \textbf{\#FLOPs} & \textbf{MMB$^{EN}$} & \textbf{MMVet}  & \textbf{TextVQA} &  \textbf{SQA$^I$} & \textbf{GQA} &  \textbf{VQAv2} & \textbf{SEED$^I$} & \textbf{POPE} & \textbf{Avg} \\
        \midrule
        \multicolumn{14}{l}{\emph{Models using Vicuna-7B}} \\
        LLaVA-1.5~\cite{llava} & Fig.~\ref{fig:pyramid}(e) & CLIP-L & 336 & 191G & 64.3 & 31.1  & 58.2 & 66.8 & 62.0 & 78.5 & 66.1 & 85.9 & 64.1 \\
        LLaVA-HR~\cite{llava-hr} & Fig.~\ref{fig:pyramid}(f) & ConvNeXt-L, CLIP-L & 1024/448 & 1172G & 66.4 & 31.2  & 67.1 & 65.1 & 64.2 & 81.9 & 64.2 & 87.6 & 66.0 \\
        \rowcolor{gray!10} PIIP-LLaVA & Fig.~\ref{fig:pyramid}(h) & CLIP-B, CLIP-L & 512/256 & 193G & 63.8 & 32.0  & 57.9 & 68.8 & 62.8 & 79.1 & 67.3 & 86.5 & 64.8 \\
        
        \rowcolor{gray!10} PIIP-LLaVA & Fig.~\ref{fig:pyramid}(h) & ConvNeXt-B, CLIP-L & 640/224 & 191G & 64.5 & 31.9 & 59.0 & 68.3 & 62.1 & 79.9 & 67.5 & 86.5 & 65.0\\
        \rowcolor{gray!10} PIIP-LLaVA & Fig.~\ref{fig:pyramid}(h) & CLIP-B, CLIP-L & 512/448 & 422G & 66.2 & 30.5 & 59.5 & 68.0 & 63.7 & 80.3 & 69.0 & 87.3 & 65.6 \\
        \rowcolor{gray!10} PIIP-LLaVA & Fig.~\ref{fig:pyramid}(h) & ConvNeXt-B, CLIP-L  & 1024/336 & 598G & 67.5 & 31.5 & 63.1 & 68.1 & 62.7 & 81.1 & 69.0 & 87.9 & 66.4\vspace{-0.6mm}\\

        \midrule
        \multicolumn{14}{l}{\emph{Models using Vicuna-13B}} \\
        LLaVA-1.5~\cite{llava} & Fig.~\ref{fig:pyramid}(e) & CLIP-L & 336 & 191G & 67.7 & 36.1  & 61.3 & 71.6 & 63.3 & 80.0 & 68.2 & 85.9 & 66.8 \\
        LLaVA-HR~\cite{llava-hr} & Fig.~\ref{fig:pyramid}(f) & ConvNeXt-L, CLIP-L & 1024/448 & 1172G & 66.5 & 34.8  & 68.1 & 68.1 & 64.8 & 82.3 & 64.5 & 87.8 & 67.1 \\
        LLaVA-UHD~\cite{llava_uhd} & Fig.~\ref{fig:pyramid}(g) & CLIP-L & 1008 & 1337G & 68.0 & $-$ & 67.7 & 72.0 & 65.2 & 81.7 & $-$ & 89.1 & $-$ \\
        \rowcolor{gray!10} PIIP-LLaVA & Fig.~\ref{fig:pyramid}(h) & CLIP-B, CLIP-L & 512/448 & 422G & 67.6 & 36.1  & 61.6 & 71.0 & 64.5 & 81.2 & 69.5 & 87.2 & 67.3 \\
        \rowcolor{gray!10} PIIP-LLaVA & Fig.~\ref{fig:pyramid}(h) & ConvNeXt-B, CLIP-L & 1024/336 & 598G & 68.5 & 37.7 & 64.2 & 71.1 & 64.2 & 81.8 & 69.3 & 87.9 & 68.0\vspace{-0.6mm}\\
        \bottomrule 
        \end{tabular}
    }
    \label{table:multimodal_baseline}
\end{table*}
\noindent
\begin{table*}[t]
    \small
    \centering
    \caption{
    \textbf{Comparison with existing MLLMs on multimodal benchmarks.} 
    ``Data'' denotes the data size of all training stages. 
    *Images from SQA-IMG test set are observed in training, so we mark its result and Avg in gray. 
    }
    \vspace{-1mm}
    \renewcommand\arraystretch{1.5}
    \resizebox{\textwidth}{!}{
        
        \begin{tabular}{c|cccc|cccccccc|c}
        \toprule
        \textbf{Method} & \textbf{Vision Encoder} & \textbf{Resolution} & \textbf{LLM} & \textbf{Data} & \textbf{MMB$^{EN}$} & \textbf{MMVet}  & \textbf{TextVQA} &  \textbf{SQA$^I$} & \textbf{GQA} &  \textbf{VQAv2} & \textbf{SEED$^I$} & \textbf{POPE} & \textbf{Avg} \\
        \midrule
        InstructBLIP~\cite{VLM:InstructBLIP} & ViT-g & 224 & Vicuna-7B & 130M & 38.3 & 26.2 & 50.1 & 60.5 & 49.2 & $-$ & $-$ & $-$ & $-$ \\
        QwenVL-Chat~\cite{bai2023qwenvl} & ViT-G & 448 & Qwen-7B & 1.4B & 60.9 & $-$ & 61.5 & 68.2 & 57.5 & 78.2 & 65.4 & $-$ & $-$ \\
        
        LLaVA-1.5~\cite{llava} & CLIP-L & 336 & Vicuna-7B & 1.2M & 64.3 & 31.1  & 58.2 & 66.8 & 62.0 & 78.5 & 66.1 & 85.9 & 64.1 \\
        LLaVA-HR~\cite{llava-hr} & ConvNeXt-L, CLIP-L & 1024/448 & Vicuna-7B & 1.2M & 66.4 & 31.2  & 67.1 & 65.1 & 64.2 & 81.9 & 64.2 & 87.6 & 66.0 \\
        \rowcolor{gray!10} PIIP-LLaVA & ConvNeXt-B, CLIP-L  & 1024/336 & Vicuna-7B & 1.2M & 67.5 & 31.5 & 63.1 & 68.1 & 62.7 & 81.1 & 69.0 & 87.9 & 66.4 \\
        \rowcolor{gray!10} PIIP-LLaVA & ConvNeXt-L, CLIP-L  & 1024/336 & Vicuna-7B & 1.2M & 67.0 & 31.4  & 67.1 & 68.3 & 63.9 & 81.5 & 69.4 & 88.2 & 67.1\vspace{-0.6mm}\\
        
        \midrule
        BLIP-2~\cite{VLP:BLIPv2} & ViT-g & 224 & Vicuna-13B & 129M & $-$ & 22.4 & 42.5 & 61.0 & 41.0 & 41.0 & 46.4 & 85.3 & $-$ \\
        InstructBLIP~\cite{VLM:InstructBLIP} & ViT-g & 224 & Vicuna-13B & 130M & 39.8 & 25.6 & 50.7 & 63.1 & 49.5 & $-$ & $-$ & 78.9 & $-$ \\
        Shikra~\cite{chen2023shikra} & CLIP-L & 224 & Vicuna-13B & 6M & 60.2 & $-$ & $-$ & $-$ & $-$ & 77.4 & $-$ & $-$ & $-$ \\
        
        LLaVA-1.5~\cite{llava} & CLIP-L & 336 & Vicuna-13B & 1.2M & 67.7 & 36.1  & 61.3 & 71.6 & 63.3 & 80.0 & 68.2 & 85.9 & 66.8 \\
        LLaVA-HR~\cite{llava-hr} & ConvNeXt-L, CLIP-L & 1024/448 & Vicuna-13B & 1.2M & 66.5 & 34.8  & 68.1 & 68.1 & 64.8 & 82.3 & 64.5 & 87.8 & 67.1 \\
        LLaVA-UHD~\cite{llava_uhd} & CLIP-L & 1008 & Vicuna-13B & 1.2M & 68.0 & $-$ & 67.7 & 72.0 & 65.2 & 81.7 & $-$ & 89.1 & $-$ \\
        
        \rowcolor{gray!10} PIIP-LLaVA & ConvNeXt-B, CLIP-L & 1024/336 & Vicuna-13B & 1.2M & 68.5 & 37.7 & 64.2 & 71.1 & 64.2 & 81.8 & 69.3 & 87.9 & 68.0 \\
        \rowcolor{gray!10} PIIP-LLaVA & ConvNeXt-L, CLIP-L & 1024/336 & Vicuna-13B & 1.2M & 66.5 & 36.8 & 69.2 & 69.8 & 65.2 & 82.5 & 70.5 & 87.6 & 68.5\vspace{-0.6mm}\\

        \midrule
        \multicolumn{14}{l}{\emph{More training data}} \\
        mPLUG-Owl2~\cite{VLM:mPLUG-Owl2} & CLIP-L & 448 & Llama-2-7B & 400M & 64.5 & 36.2 & 58.2 & 68.7 & 56.1 & 79.4 &  57.8 & $-$ & $-$ \\
        SPHINX-intern2~\cite{sphinx-x} & ConvNeXt-XXL, DINOv2-g & 448 & InternLM2-7B & 16M & 57.9 & 36.5 & $-$ & 70.4 & 56.2 & 75.5 & 68.8 & 86.9 & $-$ \\
        Slime~\cite{slime} & CLIP-L & 2016 & Vicuna-7B & 2M & 69.3 & 35.4 & 64.4 &  76.8 & 63.1 & 80.3 & $-$ & 85.4 & $-$ \\
        LLaVA-NeXT~\cite{VLM:LLaVA-NeXT} & CLIP-L & 1344 & Vicuna-7B & 1.6M & $-$ & 43.9 & 64.9 & 70.1 & 64.2 & 81.8 & 70.2 & 86.5 & $-$ \\
        LLaVA-UHD v2~\cite{llava_uhd_v2} & CLIP-L & 1008 & Vicuna-7B & 1.4M & 68.2 & $-$ & 67.6 & 71.3 & 65.4 & $-$ & 70.0 & $-$ & $-$ \\
        Mini-Gemini~\cite{VLM:MiniGemini} & CLIP-L & 768/336 & Vicuna-7B & 2.8M & 69.3 & 40.8 & 65.2 & $-$ & $-$ & $-$ & 68.9 & $-$ & $-$ \\
        MG-LLaVA~\cite{mg_llava} & \makecell{ConvNeXt-L, CLIP-L, \\ RAMPlus, OWL-ViTv2-L} & 768/336 & Vicuna-7B & 2.5M & 72.1 & 41.0 & 67.3 & 70.8 & $-$ & 80.2 & 69.4 & $-$ & $-$ \\
        InternLM-XC~\cite{xcomposer} & EVA-G & 224 & InternLM-7B & 1.1B & 74.4 & 35.2 & $-$ & $-$ & $-$ & $-$ & 66.9 & $-$ & $-$ \\
        MM1~\cite{VLM:MM1} & CLIP-L & 1792 & 7B & 1B & 72.3 & 42.1 & 72.8 & 72.6 & $-$ & 82.8 & 69.9 & 86.6 & $-$ \\

        \rowcolor{gray!10} PIIP-LLaVA$^{\dag}$ & ConvNeXt-L, CLIP-L & 1024/336 & Vicuna-7B & 2.7M & 74.5 & 44.7 & 73.0 & \textcolor{gray}{95.0*} & 62.9 & 82.3 & 72.1 & 87.5 & \textcolor{gray}{74.0*}\vspace{-0.6mm} \\
        \bottomrule 
        \end{tabular}
    }
    \label{table:multimodal_sota}
\end{table*}
\textbf{Setting.}
We use UperNet~\cite{upernet} as the basic framework to train on the ADE20K~\cite{ade20k} dataset based on MMSegmentation~\cite{mmseg2020}. We follow the settings of~\cite{liu2021swin} to train the model for 160k iterations. The batch size, initial learning rate and weight decay are 16, 4e-5 and 0.05.

\textbf{Results with base-size and large-size models.}
In Table~\ref{table:segmentation_baseline}, PIIP outperforms baseline with fewer computations. Moreover, PIIP-TSB in Table~\ref{table:segmentation} attains 51.6\% mIoU with UperNet, exceeding InternImage-B~\cite{internimage} by 1.4\%. Similarly, PIIP-SBL yields 54.3\% mIoU, an outstanding result compared to counterparts like ConvNeXt-XL~\cite{convnext} and InternImage-L~\cite{internimage}.

\textbf{Results with InternViT-6B.}
Similar to the conclusions in the object detection experiments, InternViT-6B equipped with PIIP achieves performance close to the baseline but saves about 58\% FLOPs, and finally achieves 59.65\% mIoU.

\subsection{Image Classification}
\label{sec:exp_cls}
\textbf{Results with pretrained models.}
We load the pretrained models for each branch and train the model for 20 epochs on ImageNet-1K~\cite{imagenet}. The batch size, learning rate and weight decay are 1024, 3e-5 and 0.1. The learning rate for the randomly initialized interactions is 10 times the base learning rate, \ie 3e-4. Other settings mainly follow the finetuning recipe of~\cite{deit3}. As shown in Table~\ref{table:cls_baseline}, when compared with the DeiT baseline, our PIIP-SBL reduces the computational cost by 36.7\% (39.0G vs. 61.6G) while maintaining the performance. When using a similar computational cost as the baseline models, PIIP-TSB and PIIP-SBL improve the top-1 accuracy by 0.3\% and 0.7\%, respectively.

\textbf{From-scratch pretraining.} As an preliminary attempt to extend our parameter-inverted image pyramid design to from-scratch pretraining, we design a model PIIP-B and evaluate it on ImageNet-1K. The specific configuration of PIIP-B is provided in Table~\ref{table:pretraining}(a), which is based on the principle that all branches should have similar computational costs, \ie the number of parameters is inversely proportional to the square of the resolution, while keeping the total number of parameters and FLOPs similar to ViT-B. Contrary to the experiments in Table~\ref{table:cls_baseline} where the classification heads of the original pretrained models are used, we use a branch merging module similar to the one in object detection and append a linear layer with GroupNorm~\cite{group_norm} for $\rm Proj(.)$, followed by a final LayerNorm and a linear classification head. We follow the pretraining recipe of~\cite{deit3} and change the number of epochs to 300. The result is provided in Table~\ref{table:pretraining}(b), where we observe that our three-branch model surpasses the baseline by 0.7\%,  highlighting the effectiveness of PIIP for from-scratch pretraining.

\subsection{Multimodal Understanding}
\label{sec:exp_multimodal}
\begin{figure*}[t]
    \centering
    \subfigure[Variants with different resolutions]{
        \begin{minipage}[t]{0.47\linewidth}
            \centering
            \includegraphics[width=1.0\linewidth]{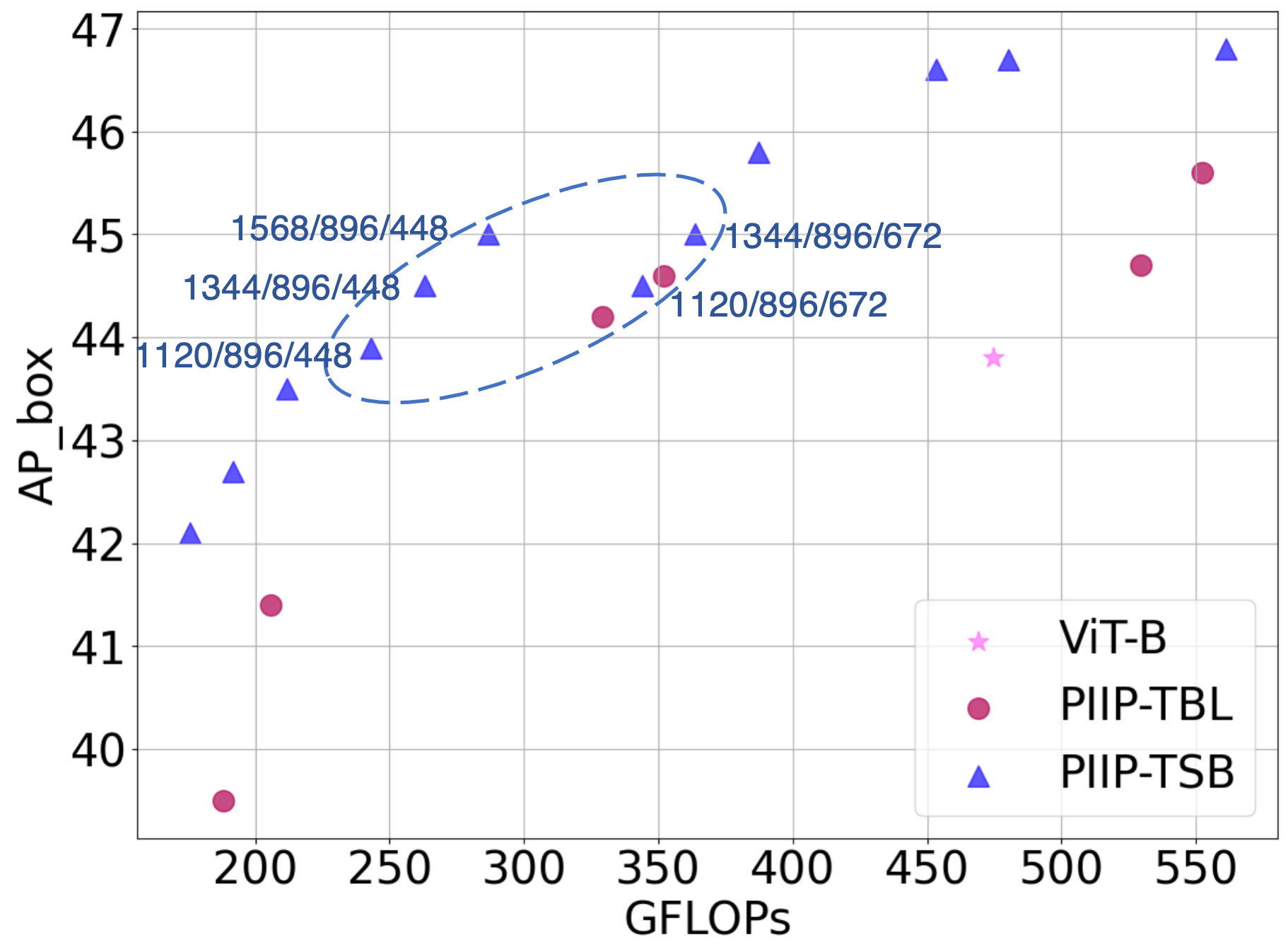}
        \end{minipage}
        \label{fig:model_variant}
    }
    \subfigure[Number of interactions]{
        \begin{minipage}[t]{0.47\linewidth}
            \centering
            \includegraphics[width=1.0\linewidth]{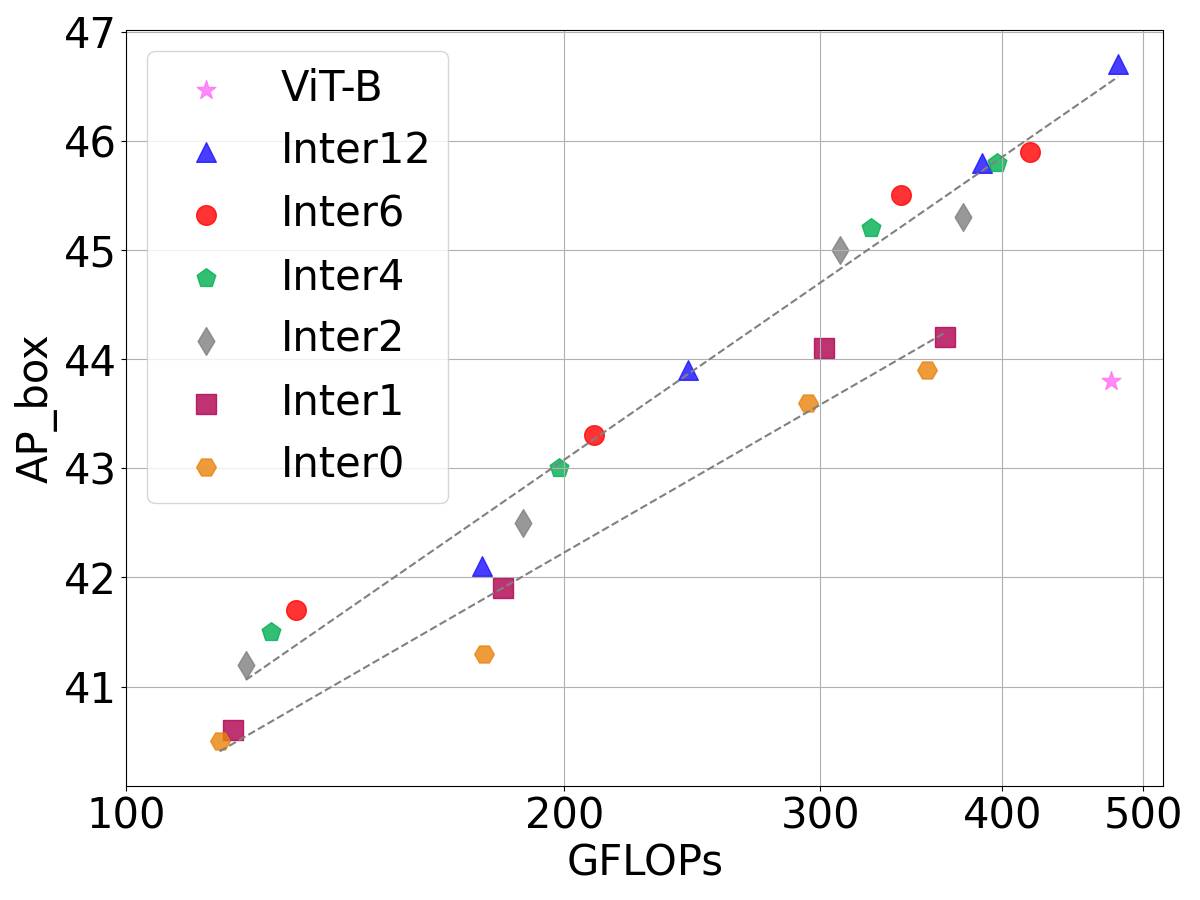}
        \end{minipage}
        \label{fig:interaction_num}
    }
    \vspace{2mm}
    \caption{\textbf{Ablation on model variants and number of interactions.}}
    \label{fig:interaction_num_whole_figure}

\end{figure*}

\textbf{Setting.} We construct our MLLM, namely PIIP-LLaVA, based on LLaVA-1.5~\cite{VLM:LLaVA-1.5}. We use a dual-branch structure with CLIP-L as Branch 1 and CLIP-B/ConvNeXt-B as Branch 2, and Vicuna-1.5-7B/13B as the language model~\cite{vicuna}, as shown in Figure~\ref{fig:multimodal}.
For the comparison with baseline (Table~\ref{table:multimodal_baseline}), we only use two interactions to balance computational cost. 
This is because we empirically find that with certain computational budget, the improvement of using more interactions is smaller in multimodal understanding (Table~\ref{table:ablation_multimodal}) than in dense prediction tasks, \ie detection and segmentation (Figure~\ref{fig:interaction_num_whole_figure}).
For the experiments in Table~\ref{table:multimodal_sota}, the total number of interactions is set to 12.
The training procedure of PIIP-LLaVA consists of two stages, \ie a pretraining stage and a instruction tuning stage, following~\cite{VLM:LLaVA-1.5}.

During pretraining, the vision encoder (branches and interactions) and language model are kept frozen, and only the projectors are optimized. Since $\gamma$ and $\tau$ in the interaction units are fixed as $\mathbf{0}$, the vision encoder is simply two frozen branches of pretrained models, and each projector serves to align the visual features of one branch with the language embedding space. We follow LLaVA-1.5 to use LCS-558K~\cite{VLM:LLaVA-1.5} for pretraining. AdamW~\cite{adamw} is used as the optimizer, and the learning rate and total batch size are set to 1e-3 and 256.

During instruction tuning, the entire MLLM is fully optimized, including two branches, interactions, projectors, and the language model. We follow LLaVA-1.5 to use 665k instruction tuning data for finetuning. The learning rate and total batch size are set to 2e-5 and 128, respectively. The training epoch is set to 1 for both the pretraining and instruction tuning stages.

When fairly comparing with recent MLLMs like MM1~\cite{VLM:MM1} which are trained with larger data scale, we incorporate additional 1.6M instruction data, including ShareGPT4V~\cite{sharegpt4v}, LAION-GPT4V~\cite{laion_gpt4v}, ALLAVA~\cite{allava}, LIMA~\cite{lima}, OpenAssistant2~\cite{openassistant}, Tabmwp~\cite{tablemwp}, MathQA~\cite{mathqa}, KVQA~\cite{kvqa},
Geometry~\cite{inter_gps}, STVQA~\cite{stvqa}, ChartQA~\cite{chartqa}, DVQA~\cite{dvqa}, AI2D~\cite{Datasets:AI2D}, LLaVA-Med~\cite{llava_med}, InfoVQA~\cite{infovqa}, MathV360k~\cite{math_llava}, and SQA~\cite{Datasets:ScienceQA}.

\textbf{Benchmarks.} We compare PIIP-LLaVA with other methods on 8 multimodal benchmarks: MMBench-EN~\cite{Datasets:MMBench}, MMVet~\cite{Datasets:MM-vet}, TextVQA~\cite{Datasets:TextVQA}, SQA-IMG~\cite{Datasets:ScienceQA}, GQA~\cite{Datasets:GQA}, VQAv2~\cite{Datasets:VQAv2}, SEED Image~\cite{Datasets:Seed-bench}, and POPE~\cite{Datasets:POPE}. 
In particular, MMBench, MMVet and SEED Image evaluate the multimodal perception and cognition abilities of
MLLMs. POPE measures the visual hallucinations of MLLMs. TextVQA, SQA-IMG, GQA and VQAv2 evaluate the general visual question answering capability. Specifically, TextVQA contains text-rich images for fine-grained recognition, which requires high-resolution understanding capability.

\textbf{Comparison with multi-resolution baselines.} Table~\ref{table:multimodal_baseline} shows the comparison with other multi-resolution baselines. We observe that with nearly equivalent computation costs, PIIP-LLaVA surpasses LLaVA-1.5~\cite{VLM:LLaVA-1.5} on all of the benchmarks, with +1.1\% on average and +2.0\% on SQA-Image. This is because the parameter-inverted design helps to build efficient image pyramids, thereby allowing for larger input resolution (\ie 512 or 640 compared with 336 of LLaVA-1.5) and stronger image understanding ability.
 We also find that heterogeneous structures (ConvNeXt-B + CLIP-L) show advantages when handling high resolution inputs, as CNN contribute to extracting local details and ViT extract high-level semantics, similar to the analysis in Section~\ref{sec:method_branches}.
 
 When we scale up the input resolution and compare PIIP-LLaVA with image pyramid method LLaVA-HR~\cite{llava-hr} and dynamic high resolution method LLaVA-UHD~\cite{llava_uhd}, PIIP-LLaVA achieves comparable or better performance with 44\% or 51\% of the computational costs.
 These results confirm the generalization of PIIP as an efficient and effective multi-resolution approach in multimodal understanding tasks.

\textbf{Comparison with existing MLLMs.} In Table~\ref{table:multimodal_sota}, we further scale up the ConvNeXt model and the training data size and compare PIIP-LLaVA with existing MLLMs. When trained with 1.2M data, PIIP-LLaVA achieves superior performance over existing MLLMs, \eg +1.4\% over LLaVA-HR-13B. When trained with more data, PIIP-LLaVA outperforms all models, including MM1~\cite{VLM:MM1} with 1B training data and 1792 resolution, \eg 74.5\% on MMBenc, 73.0\% on TextVQA, and 87.5\% on POPE. These results demonstrate the effectiveness of PIIP on multimodal perception and cognition, general visual question answering and eliminating hallucination.
\noindent
\begin{table*}[!h]
    \small
    \centering
    \caption{\textbf{Ablation on multimodal understanding with CLIP-B, CLIP-L and Vicuna-7B.}}
    \vspace{-1mm}
    \renewcommand\arraystretch{1.4}
    \resizebox{\textwidth}{!}{
        
        \begin{tabular}{cccc|cccccccc|c}
        \toprule
        \textbf{\#Interactions} & \textbf{Pretrain Unfreeze} & \textbf{Resolution} & \textbf{\#FLOPs} & \textbf{MMB$^{EN}$} & \textbf{MMVet}  & \textbf{TextVQA} &  \textbf{SQA$^I$} & \textbf{GQA} &  \textbf{VQAv2} & \textbf{SEED$^I$} & \textbf{POPE} & \textbf{Avg} \\
        \midrule
        2 & none & 512/256 & 193G & 65.0 & 34.0 & 57.9 & 68.8 & 62.8 & 79.1 & 67.3 & 86.5 & 65.2 \\
        0 & none & 528/256 & 196G & 65.7 & 29.5 & 56.8 & 69.2 & 63.0 & 78.9 & 66.8 & 85.4 & 64.4 \\
        12 & none & 464/256 & 194G & 64.6 & 31.3 & 57.4 & 69.5 & 62.7 & 79.1 & 67.1 & 86.1 & 64.7 \\
        2 & interactions & 512/256 & 193G & 63.2 & 25.4 & 52.4 & 70.6 & 61.1 & 75.9 & 62.7 & 85.5 & 62.1 \\
        2 & all & 512/256 & 193G & 34.3 & 15.1 & 43.3 & 67.7 & 44.6 & 52.3 & 41.0 & 71.9 & 46.3 \\
        \bottomrule 
        \end{tabular}
    }
    \label{table:ablation_multimodal}
\end{table*}
\begin{table*}[!t]
\centering
\caption{\textbf{Ablation on image pyramid and parameter-inverted design.} `PI', `IP' and `Inter.' represent parameter-inverted, image pyramid and interactions. `MS' means multi-scale training, following~\cite{autoaugment}.
}
\renewcommand\arraystretch{1.3}
\resizebox{\linewidth}{!}{
    \begin{tabular}{c|ccccc|cc|cccccc}
    \toprule
    \multirow{2}{*}{\textbf{Figure}} & \multirow{2}{*}{\textbf{Branches}} & \multirow{2}{*}{\textbf{PI}} & \multirow{2}{*}{\textbf{IP}} & \multirow{2}{*}{\textbf{Inter.}} & \multirow{2}{*}{\textbf{Resolution}} & \multirow{2}{*}{\textbf{\#Param}} & \multirow{2}{*}{\textbf{\#FLOPs}} & \multicolumn{6}{c}{\textbf{Mask R-CNN 1$\times$ schedule}} \\ 
    & & & & & & & & $\rm AP^b$ & $\rm AP^b_{50}$ & $\rm AP^b_{75}$ & $\rm AP^m$ & $\rm AP^m_{50}$ & $\rm AP^m_{75}$ \\
    \midrule
    Fig.~\ref{fig:pyramid}(a) & B &    &   &   & 1024 & 90M & 463G & 43.8 & 67.6 & 47.7 & 39.9 & 63.6 & 42.2  \\
    Fig.~\ref{fig:pyramid}(b) & B &    & \checkmark &   & MS & 90M & 463G & 44.8 & 69.2 & 49.1 & 41.0 & 65.8 & 43.9 \\
    - & BBB &    & \checkmark &   & 896/448/224 & 262M & 369G & 43.3 & 65.8 & 46.6 & 37.9 & 61.5 & 39.6 \\
    - & BBB &    & \checkmark &   & 896/672/224 & 263M & 457G & 43.8 & 66.3 & 47.3 & 38.2 & 62.2 & 39.7 \\
    Fig.~\ref{fig:pyramid}(c) & BBB &    & \checkmark & \checkmark & 896/448/224 & 341M & 466G & 44.5 & 66.5 & 48.2 & 38.7 & 62.6 & 40.6 \\
    - & TSB &    &   & \checkmark & 896/896/896 & 148M & 468G & 44.6 & 66.4 & 48.3 & 39.0 & 62.7 & 41.4 \\
    Fig.~\ref{fig:pyramid}(d) & TSB &    & \checkmark & \checkmark & 448/672/896 & 147M & 452G & 42.6 & 64.2 & 45.6 & 36.5 & 59.5 & 38.0 \\
    \rowcolor{gray!10} Fig.~\ref{fig:pyramid}(h) & TSB & \checkmark & \checkmark & \checkmark & 1568/1120/672 & 149M & 453G & \textbf{46.6} & 68.4 & 51.1 & \textbf{41.4} & 65.2 & 44.3\vspace{-0.6mm} \\
    
    \midrule
    
    Fig.~\ref{fig:pyramid}(a) & L &   &   &   & 1024 & 308M & 1542G & 46.8 & 70.8 & 51.4 & 42.5 & 67.3 & 45.3 \\
    Fig.~\ref{fig:pyramid}(c) & LLL &   & \checkmark & \checkmark & 896/448/224 & 1053M & 1458G & 46.9 & 69.7 & 51.2 & 40.8 & 65.3 & 43.3 \\
    - & SBL &   &   & \checkmark & 848/848/848 & 495M & 1539G & 47.2 & 69.4 & 51.0 & 41.1 & 65.4 & 43.7 \\
    \rowcolor{gray!10} Fig.~\ref{fig:pyramid}(h) & SBL & \checkmark & \checkmark & \checkmark & 1568/896/672 & 497M & 1464G & \textbf{49.4} & 71.9 & 53.9 & \textbf{43.7} & 68.4 & 46.6\vspace{-0.6mm} \\
    \bottomrule
    \end{tabular}
}
\label{table:image_pyramid}
\end{table*}
\begin{table}[t]
\small
\centering
\caption{Baseline with higher resolution.}
\renewcommand\arraystretch{1.4}
\resizebox{\columnwidth}{!}{
    \begin{tabular}{c|cccc}
    \toprule
    \textbf{Model} & \textbf{Resolution} & \textbf{\#Param} & \textbf{\#FLOPs} & \textbf{$\rm AP^b$} \\
    \midrule
    ViTDet-L &	1024 & 308M & 1542G & 46.8 \\
    ViTDet-L & 1792 & 308M & 6458G & 48.3 \\
    PIIP-TSBL & 1792/1568/1120/448 & 512M & 1535G & \textbf{49.4} \\
    \bottomrule
    \end{tabular}
}
\label{table:large_res}
\end{table}
\begin{table*}[]
\small
\centering
\caption{\textbf{Ablation on attention implementation and number of interactions} with PIIP-TSB 1120/896/448. }
\vspace{1mm}
\renewcommand\arraystretch{1.2}
\resizebox{0.85\linewidth}{!}{
    \begin{tabular}{c|ccccc|ccccc}
    \toprule
    \multirow{2}{*}{\textbf{\#Interaction}} & \multicolumn{5}{c|}{\textbf{Regular Attention}} & \multicolumn{5}{c}{\textbf{Deformable Attention}} \\ 
    & \#FLOPs & $\rm AP^b$ & $\rm AP^b_{l}$ & $\rm AP^b_{m}$ & $\rm AP^b_{s}$ & \#FLOPs & $\rm AP^b$ & $\rm AP^b_{l}$ & $\rm AP^b_{m}$ & $\rm AP^b_{s}$ \\
    \midrule
    0 & 176G & 41.3 & 59.0 & 44.6 & 22.5  & 176G & 41.3 & 59.0 & 44.6 & 22.5 \\
    1 & 211G & 41.1 & 59.1 & 44.9 & 22.6 & 182G & 41.9 & 59.8 & 45.5 & 22.4 \\
    2 & 245G & 41.7 & 59.5 & 45.2 & 22.7 & 187G & 42.5 & 60.5 & 46.4 & 23.1 \\
    4 & 315G & 41.6 & 59.2 & 45.3 & 22.8 & 198G & 43.0 & 61.0 & 47.3 & 23.3 \\
    6 & 384G & 42.1 & 59.7 & 45.8 & 23.2 & 210G & 43.3 & 61.8 & 46.9 & 23.6 \\
    12 & 592G & 42.0 & 60.0 & 45.9 & 23.1 & 243G & \textbf{43.9} & 62.4 & 47.9 & 24.4 \\
    \bottomrule
    \end{tabular}
}
\label{table:interaction}
\end{table*}
\subsection{Ablation Study}
\label{sec:exp_ablation}
We conduct ablation studies to evaluate the impact of different design choices. We first study the influence of unfreezed modules and number of interactions on multimodal understanding, and then provide extensive experimental results on MS COCO object detection. 

\textbf{Ablation on multimodal understanding tasks.} As shown in Table~\ref{table:ablation_multimodal}, we test different numbers of interactions and unfreezed modules of the vision encoder during pretraining, and adjust the resolution to maintain computational cost unchanged. We observe that while 2 interactions achieve better performance than no interactions, 12 interactions result in negative effect. For unfreezed modules, when unfreeze the interactions or the entire vision encoder (branches and interactions) during pretraining, the performance declines significantly. This suggests the importance of training solely the projectors and aligning the vision encoder with the language model.

\textbf{Superiority of parameter-inverted image pyramid networks.}
In Table~\ref{table:image_pyramid}, we evaluate the effectiveness of the image pyramid and parameter-inverted design by comparing our method with other methods, \eg designs in Figure~\ref{fig:pyramid}. We provide three analyses:
\textbf{1)} A single-branch model with multi-scale training represents the most basic implementation of an image pyramid (row 2). Compared with the baseline model, its performance improvement is limited (44.8\% vs. 43.8\%). 
\textbf{2)} Interactions provide cross-branch feature fusion that are beneficial for visual perception (row 3,4,5).
\textbf{3)} We conduct experiments by controlling the scale of the branches and the input resolutions while keeping the computational cost close. Specifically, when using the same input image resolution, the combination of models of different sizes does not bring significant improvements to detection performance (row 6). Correspondingly, when the three branches use the same model (\eg BBB), the input image resolution is adjusted to the pyramid structure  (row 5). The performance of the final model is slightly improved on $\rm AP^b$ (44.5\% vs. 43.8\%), but the $\rm AP^m$ drops significantly (38.7\% vs 39.9\%) due to the reduction of the maximum resolution. The former demonstrates the importance of the image pyramid, and the latter further demonstrates the need for the image pyramid to maintain a larger image scale range, which is especially essential for the instance segmentation task. 

Drawing on experience, parameter-inverted image pyramid networks are an efficient design that meets the above requirements, especially when compared to its opposite configuration parameter-direct image pyramid (row 7), \ie TSB with 448/672/896 resolution (46.6\% vs. 42.6\%). With less computation than the baseline, PIIP can support image inputs in the maximum range from 672 to 1568 (row 8), and the performance is significantly improved.

\textbf{Baseline with higher resolution.}
To evaluate the impact of using higher resolutions, we add another baseline ViTDet-L with 1792 resolution (matching the largest resolution of PIIP-TSBL 1792/1568/1120/448), as shown in the second row of the Table~\ref{table:large_res}. The first and third rows are from Table 1. We observe that while ViTDet-L with 1792 resolution achieves better performance compared to the 1024 resolution, its FLOPs are approximately 4 times larger. Compared with PIIP-TSBL, ViTDet-L 1792 has a lower box AP (-1.3\%) and incurs 4 times the computational cost in FLOPs. This experiment explains that the performance improvement does not entirely come from larger image resolution. The structure of PIIP also makes an important contribution in the computational cost and performance improvement.

\textbf{Design guidelines for PIIP.}
Through extensive practice, we derive empirical design guidelines when scaling up the model:
1) Prioritize increasing the image resolution of the largest image branch: as shown in the blue dashed circle in Figure~\ref{fig:interaction_num_whole_figure} (a), the input resolution of the largest image branch is greatly increased without causing a sharp increase in the total computational cost. 2) The scale of the largest model does not need to exceed the baseline model: the introduction of larger models restricts the resolution range of the image pyramid, \eg PIIP-TSB is more cost-effective than PIIP-TBL according to Figure~\ref{fig:interaction_num_whole_figure} (a).

\textbf{Branch merging.}
We examine the effects of using feature maps from all branches in branch merging. Experiments in Table~\ref{table:merge} demonstrate that merging all branches yields the best performance by providing multi-scale semantically rich features, compared to using feature maps from only one or two of the branches. 

\textbf{Attention implementation.}
The core of information interaction between branches is cross-attention mechanism. We adopt PIIP-TSB with resolution 1120/896/448 as the basic model and investigate two different attention implementations. As shown in Table~\ref{table:interaction}, deformable attention~\cite{deformable_detr} with linear complexity can significantly improve the performance without substantially increasing the computational cost. We opt for using deformable attention as the default configuration.

\textbf{Number of interactions.}
We test different number of interactions in object detection. As shown in Table~\ref{table:interaction}, regardless of the attention implementations, the increase in the number of interactions helps to improve the performance to varying degrees. 
Since this also increases the computational cost, we further explore the cost-effectiveness of the number of interactions. We conduct experiments with multiple resolution combinations on models with different numbers of interactions, and the scatter plot of all results is shown in Figure~\ref{fig:interaction_num_whole_figure} (b). 

It can be seen that when the number of interactions is small (\ie less than 2), the growth trend of model performance with the increase in computational cost is relatively slow. When the number of interactions is 4, 6, or 12, the trend is similar. This is different from our observation in multimodal understanding. We attribute this to the fact that object detection requires higher resolution (\eg detecting small objects) and therefore more cross-scale feature fusion. For multimodal understanding, the resolution of all branches are closer, and the task itself focuses more on semantics rather than detailed localization.

\begin{figure}
    \centering
    \includegraphics[width=0.9\linewidth]{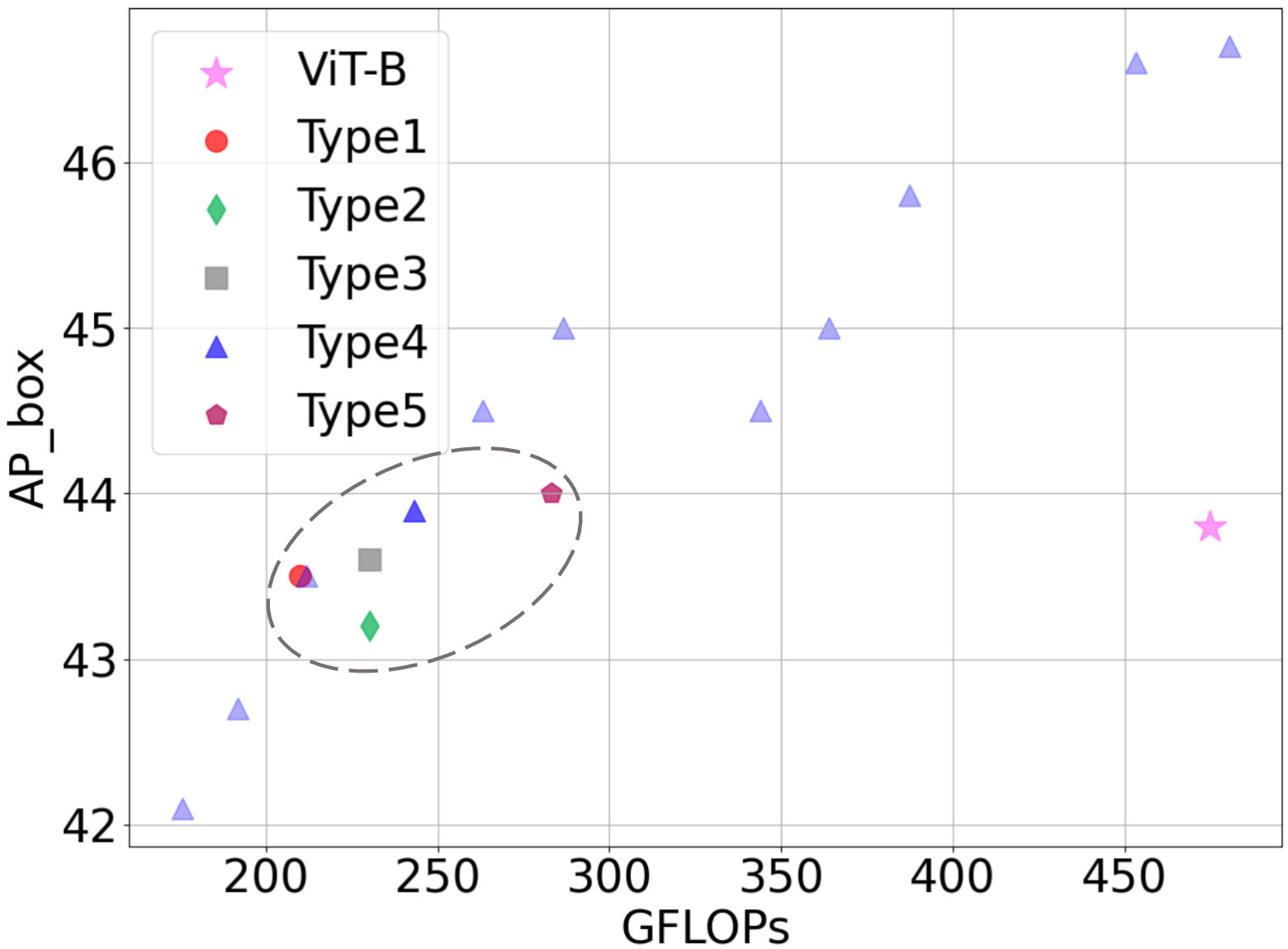}
    \caption{\textbf{Performance of different interaction directions.}}
    \label{fig:interaction_type_scatter}
\end{figure}

\textbf{Interaction direction between branches.}
We compare five different interaction directions in Table~\ref{table:interaction_type}. 
Considering both the computational cost and performance, 
we ultimately select the fourth method, \ie bidirectional connections of adjacent branches, as the default choice. 
As can be seen from Figure~\ref{fig:interaction_type_scatter}, 
all the interaction directions achieve a satisfactory performance-computation balance, validating their ability to improve communication between branches. 

\subsection{Visualization}
\label{sec:exp_visualization}

\begin{figure}[t]
  \centering
   \includegraphics[width=\linewidth]{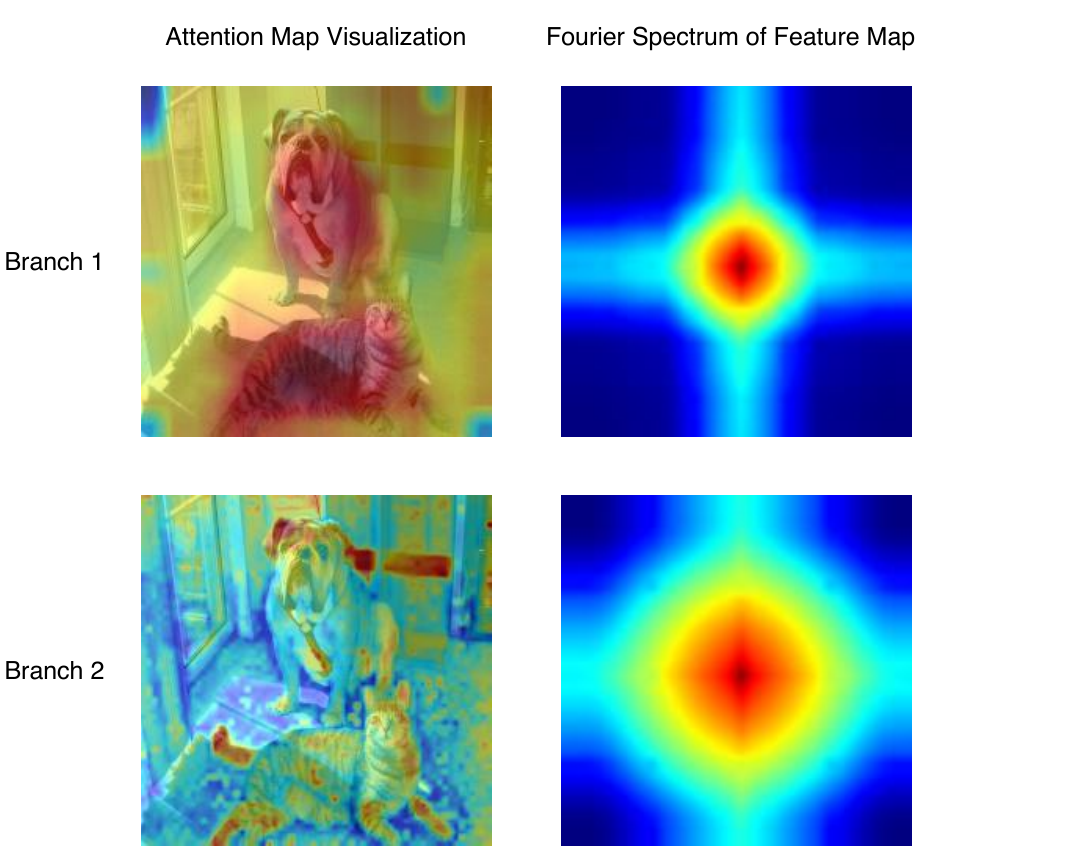}
   \caption{\textbf{Attention map visualization and Fourier spectrum of feature maps from two branches.} Branch 1 with higher resolution focuses on the semantics-rich objects of low-frequency components. Branch 2 with lower resolution highlights localization like edges and details of high-frequency components.
   }
   \label{fig:attn_map}
\end{figure}

To gain a deeper understanding of the internal mechanisms of PIIP, we visualize the attention map and Fourier spectrum of feature maps of a two-branch model, \ie Block 6 of PIIP-H6B in object detection. As shown in Figure~\ref{fig:attn_map}, the two branches display distinct patterns. Branch 1, the low-resolution one, focuses on the global structure and semantic information. Its attention map highlights key objects in the image, like the dog and cat in the example. The Fourier spectrum demonstrates that feature map of Branch 1 contains more low-frequency components. 

On the contrary, Branch 2 with high resolution concentrates on details of the image with localization information. The attention map emphasizes edges of the objects, and the feature map has more high-frequency components. These findings are consistent with our motivation in Section~\ref{sec:intro} that the low-resolution branch extracts global semantics and the high-resolution branch captures detailed information.

\subsection{Qualitative Analysis}
\label{sec:exp_qualitative}

\begin{figure*}[h]
  \centering
  \includegraphics[width=\linewidth]{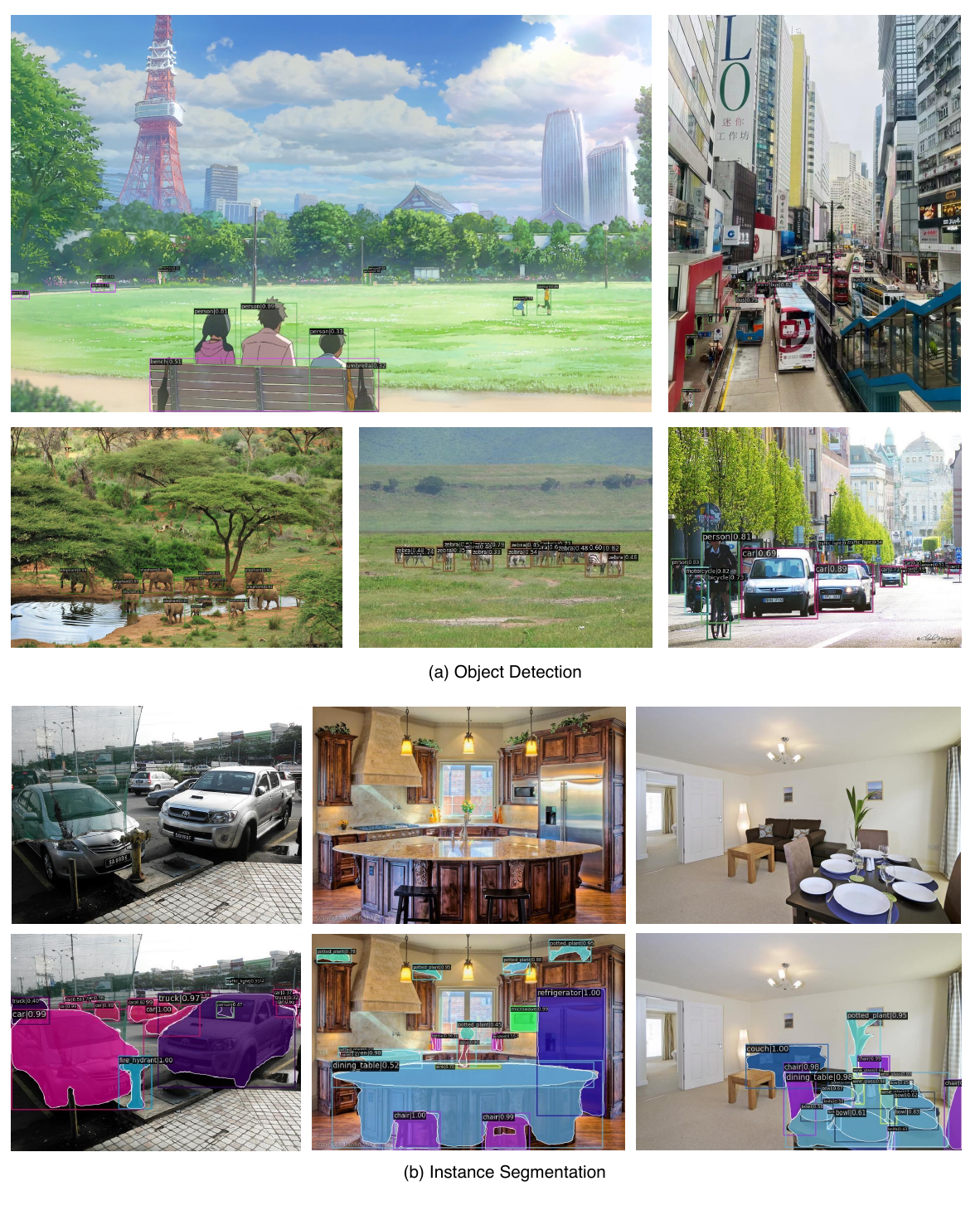}
  \caption{\textbf{Qualitative results on object detection and instance segmentation.} Please zoom in to view detailed bounding boxes and masks. High-resolution processing capability enables PIIP to accurately detect small objects in the images.
  }
  \label{fig:det_qualitative}
\end{figure*}

Figure~\ref{fig:det_qualitative} shows examples of PIIP-SBL on object detection and instance segmentation. Owing to the high-resolution processing capability, our model can detect small objects accurately, like the tiny bench and persons in the background of the upper-left image, or the small cars in the upper-right image, demonstrating its ability in visual perception.

\begin{figure*}[!t]
  \centering
  \includegraphics[width=\linewidth]{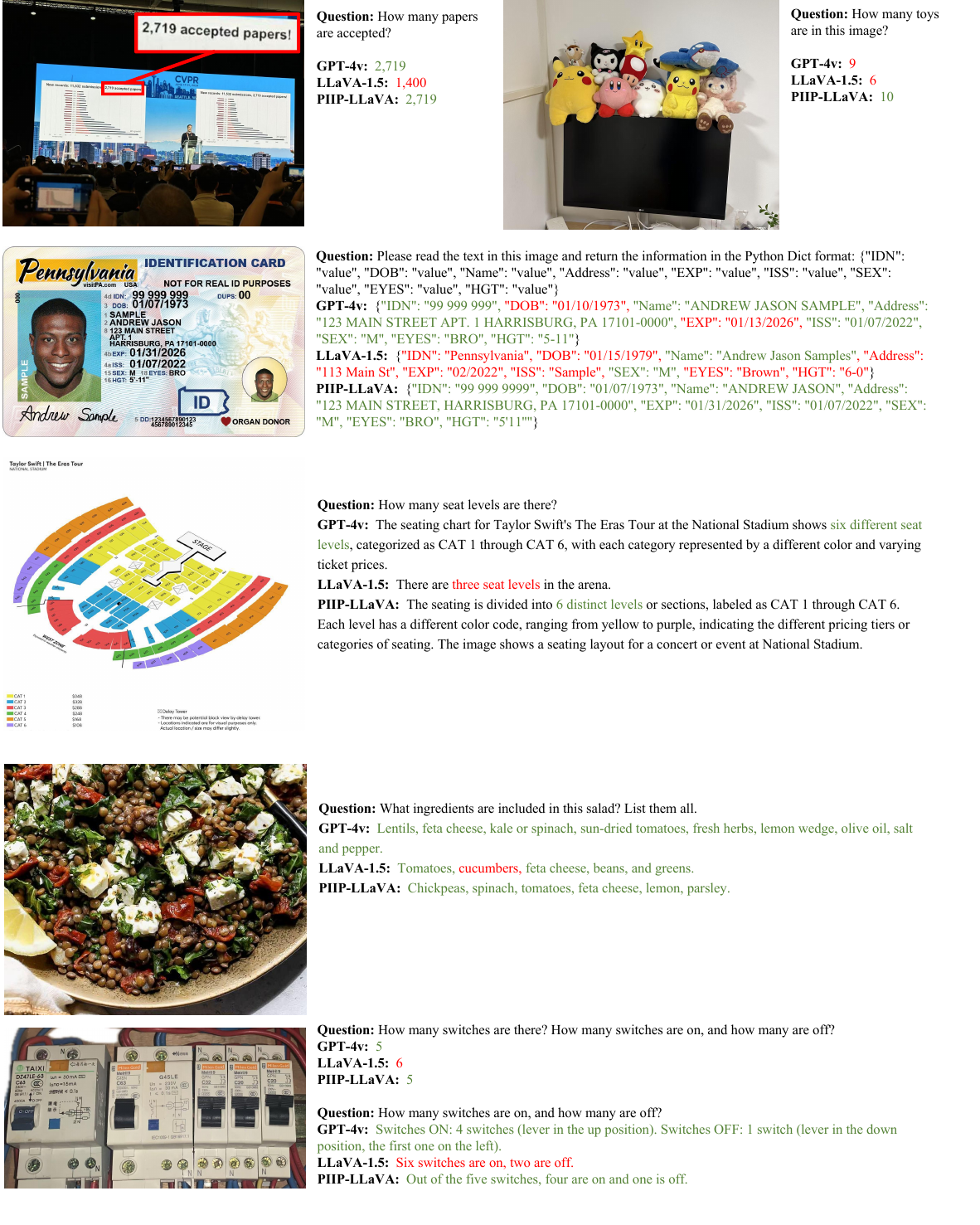}
  \caption{\textbf{Qualitative results of PIIP-LLaVA on multimodal understanding.} Red denotes incorrect answers and green denotes correct ones. PIIP-LLaVA is capable of tackling fine-grained vision-language tasks like counting, text recognition and visual information extraction. 
  }
  \label{fig:mm_qualitative}
\end{figure*}

In Figure~\ref{fig:mm_qualitative}, we provide qualitative results of PIIP-LLaVA in multimodal understanding. Compared with the LLaVA-1.5 baseline, PIIP-LLaVA has stronger ability on fine-grained vision-language tasks like text recognition (row 1 left and 3), counting (row 1 right and 5), and fine-grained recognition (row 4). 
For the challenging task visual information extraction (row 2), PIIP-LLaVA can extract the visual information in the identification card and organize it correctly, which is even comparable with GPT-4v~\cite{VLM:GPT-4v}.
These results highlight the effectiveness of PIIP in multimodal understanding.

\begin{table*}[!t]
\centering\caption{\textbf{Ablation on interaction directions} with PIIP-TSB under resolution 1120/896/448.}
\vspace{2mm}
\renewcommand\arraystretch{1.3}
\resizebox{0.65\linewidth}{!}{
        
        \begin{tabular}{c|ccccc}
        \toprule
        \textbf{Type} & 
        
        \vspace{1.5mm}
        
        \begin{minipage}[b]{0.12\columnwidth}
    	\centering
    	\raisebox{-.5\height}{\includegraphics[width=0.98\linewidth]{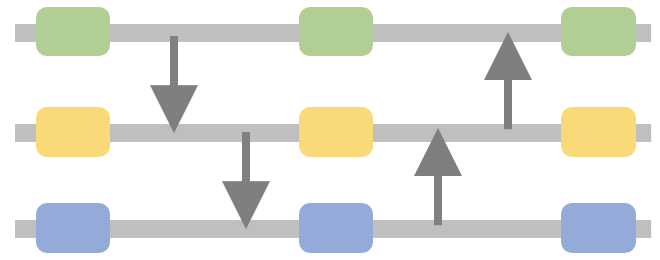}}
        \end{minipage} &
        \begin{minipage}[b]{0.12\columnwidth}
    	\centering
    	\raisebox{-.5\height}{\includegraphics[width=0.98\linewidth]{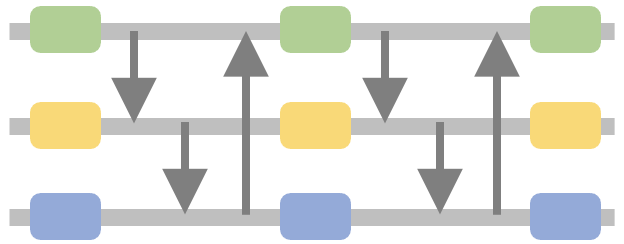}}
        \end{minipage} &
        \begin{minipage}[b]{0.12\columnwidth}
    	\centering
    	\raisebox{-.5\height}{\includegraphics[width=0.98\linewidth]{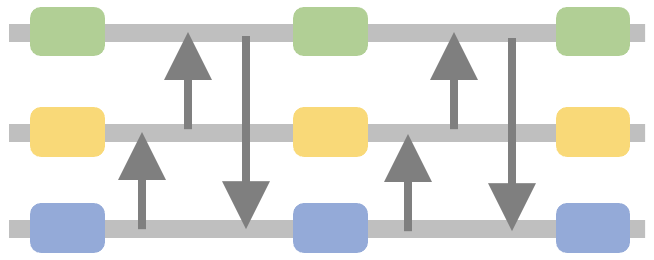}}
        \end{minipage} & 
        \begin{minipage}[b]{0.12\columnwidth}
    	\centering
    	\raisebox{-.5\height}{\includegraphics[width=0.98\linewidth]{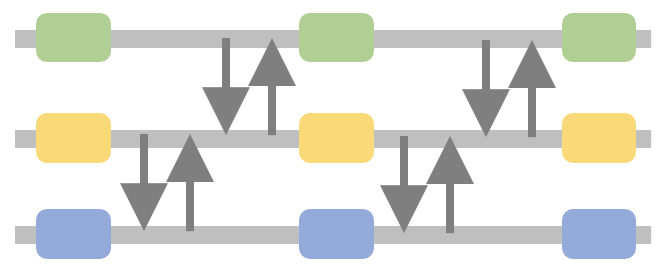}}
        \end{minipage} &
        \begin{minipage}[b]{0.1\columnwidth}
    	\centering
    	\raisebox{-.5\height}{\includegraphics[width=0.98\linewidth]{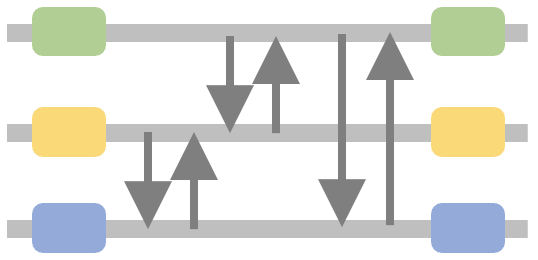}}
        \end{minipage}

        \\
        \cmidrule{1-6}
        \#FLOPs & 210G & 230G & 230G & 243G & 283G \\
        \textbf{$\rm AP^b$} & 43.5 & 43.2 & 43.6 & 43.9 & 44.0 \\
        \textbf{$\rm AP^m$} & 38.7 & 38.3 & 38.6 & 38.6 & 38.7 \\
        \bottomrule
    
        \end{tabular}
}

\label{table:interaction_type}
\end{table*}
\vspace{1mm}

\section{Conclusion}
In this paper, we propose the Parameter-Inverted Image Pyramid Networks (PIIP), a novel architecture that addresses the computational inefficiencies of traditional image pyramid methods by pairing smaller models with high-resolution images and larger models with low-resolution images. PIIP leverages pretrained models, heterogeneous ViT-CNN architectures, and a dedicated feature interaction mechanism. Through extensive experiments on visual perception and multimodal understanding tasks, such as object detection, segmentation, image classification, and multimodal understanding, PIIP demonstrates its effectiveness in balancing computational cost and performance. It achieves superior results over single-branch and existing multi-resolution approaches, achieving state-of-the-art performance on large-scale vision foundation models and multimodal large language models while significantly reducing computational requirements.

Our contributions validate the versatility of PIIP in addressing diverse challenges in vision perception and understanding. 
We hope this work inspires further exploration into efficient and effective multi-scale feature extraction for future vision and multimodal computing tasks.

\clearpage

\clearpage

\clearpage

\small{
    \bibliographystyle{IEEEtran}
    \bibliography{main}
}

\vfill

\end{document}